\documentclass[twoside,11pt]{article}

\usepackage[abbrvbib, preprint]{jmlr2e}
\usepackage{amsmath}
\usepackage{bm}
\usepackage{algorithmic}
\usepackage{threeparttable}
\usepackage{booktabs}
\usepackage{placeins}
\usepackage{color}
\usepackage{siunitx}
\usepackage[all]{nowidow}

\allowdisplaybreaks

\usepackage[ruled, noend, noline, vlined]{algorithm2e}
\makeatletter
    \newenvironment{Ualgorithm}[1][htpb]{\def\@algocf@post@ruled{\kern\interspacealgoruled\hrule  height\algoheightrule\kern3pt\relax}%
    \DontPrintSemicolon
    \begin{algorithm}[#1]}
{\end{algorithm}}
\makeatother
\SetKwComment{Comment}{$\triangleright$\ }{}
\SetKw{KwNot}{not}
\newcommand{\BlankLineA}{\vspace{2.5mm}}

\usepackage[font=small,skip=10pt]{caption}

\newcommand{\figref}[2]{\hyperref[{#1}]{\ref*{#1}#2}}
\newcommand{\opl}[1]{#1 & \;}
\newcommand{\opr}[1]{& \; #1}

\newcommand{\M}{\mathbf{M}}
\newcommand{\T}{^{\top}}
\newcommand{\KE}{\kappa_\epsilon}
\newcommand{\KP}{\kappa_{\text{pen}}}
\newcommand{\UWU}{\mathbf{U}\T\mathbf{W}\mathbf{U}}
\newcommand{\uWu}[2]{\mathbf{u}_{#1}\T\mathbf{Wu}_{#2}}
\DeclareMathOperator*{\tr}{tr}
\DeclareMathOperator*{\median}{median}
\DeclareMathOperator*{\diag}{diag}
\DeclareMathOperator*{\mean}{mean}
\DeclareMathOperator*{\argmin}{arg\,min}

\newcommand\bigO[1]{\ensuremath{\mathcal{O}({#1})}}

\newcommand{\AlgStep}[2]{Step~\hyperref[#1]{#2}}

\defcitealias{leeuw1977IM}{De Leeuw}
\defcitealias{leeuw1988convergence}{De Leeuw}
\defcitealias{leeuw1994mm_convergence}{De Leeuw}
\defcitealias{deleeuw1980step_doubling}{De Leeuw and Heiser}

\jmlrheading{volume}{2023}{pages}{date submitted}{date published}{paper id}{Daniel J.W. Touw, Patrick J.F. Groenen, and Yoshikazu Terada}

\ShortHeadings{Convex Clustering through MM (CCMM)}{Touw, Groenen, and Terada}
\firstpageno{1}

\begin{document}

\title{Convex Clustering through MM: An Efficient Algorithm to Perform Hierarchical Clustering}

\author{\name Daniel J.W. Touw \email touw@ese.eur.nl \\
       \name Patrick J.F.\ Groenen \email groenen@ese.eur.nl \\
       \addr Econometric Institute\\
       Erasmus University Rotterdam\\
       P.O. Box 1738\\
       3000 DR Rotterdam\\
       The Netherlands
       \AND
       \name Yoshikazu Terada \email terada.yoshikazu.es@osaka-u.ac.jp\\
       \addr Graduate School of Engineering Science\\
       Osaka University\\
       1-3 Machikaneyama-cho, Toyonaka\\
       Osaka 560-8531, Japan
}
\editor{?}

\maketitle

\begin{abstract}%   <- trailing '%' for backward compatibility of .sty file
Convex clustering is a modern method with both hierarchical and $k$-means clustering characteristics.
Although convex clustering can capture complex clustering structures hidden in data, the existing convex clustering algorithms are not scalable to large data sets with sample sizes greater than several thousands.
Moreover, it is known that convex clustering sometimes fails to produce a complete hierarchical clustering structure.
This issue arises if clusters split up or the minimum number of possible clusters is larger than the desired number of clusters.
In this paper, we propose convex clustering through majorization-minimization (CCMM)---an iterative algorithm that uses cluster fusions and a highly efficient updating scheme derived using diagonal majorization.
Additionally, we explore different strategies to ensure that the hierarchical clustering structure terminates in a single cluster.
With a current desktop computer, CCMM efficiently solves convex clustering problems featuring over one million objects in seven-dimensional space, achieving a solution time of 51 seconds on average.
\end{abstract}


\section{Introduction}
Clustering is a core method in unsupervised machine learning and is used in many fields. Applications range from customer segmentation in marketing analytics to object detection in image processing. Over the past two decades, a new method called convex clustering has been introduced that combines aspects of the two popular techniques: $k$-means \citep{macqueen1967kmeans} and hierarchical clustering \citep[see, for example,][]{ma2007clustering_book}.

The convex clustering framework was introduced by \cite{pelckmans2005convex_first} who formulated clustering as shrinkage of centroid distances subject to a convex loss function where each $p$-dimensional object $\mathbf{x}_i$ is represented by its own centroid $\mathbf{a}_i$.
Distance shrinkage is applied to the pairs of $\mathbf{a}_i$ and $\mathbf{a}_j$ using a penalty term, and clusters emerge via the occurrence of equal cluster centroids. \cite{hocking2011clusterpath} have explained convex clustering as a convex relaxation of hierarchical clustering, while \cite{lindsten2011convex_also} have staged it as a convex relaxation of $k$-means clustering. 
Furthermore, early explorations into variants of the standard convex clustering problem can be found in \cite{she2008sparse} and \cite{guo2010pairwise}.

The specific loss function characterizing the convex clustering model is
\begin{equation}
\label{eq:clusterloss_classic}
    L(\mathbf{A}) = \frac{1}{2} \| \mathbf{X - A} \|^2 + \lambda \sum_{i<j} w_{ij} \|\mathbf{a}_i - \mathbf{a}_j\|,
\end{equation}
where $\mathbf{A}$ is a $n \times p$ representation matrix whose rows $\mathbf{a}_i^\top$ are centroids corresponding to objects,
$w_{ij}$ is a user-defined nonnegative weight that reflects the importance of clustering objects $i$ and $j$, $\mathbf{X}$ is the data matrix with rows $\mathbf{x}_i^\top$, and $\lambda$ is a penalty-strength parameter that controls the shrinkage of cluster-centroids. Our notation $\| \cdot \|$ denotes either the $\ell_2$-norm of a vector or the Frobenius norm of a matrix.
The second term consists of $\|\mathbf{a}_i - \mathbf{a}_j\|$ representing the Euclidean distance between rows $i$ and $j$ of $\mathbf{A}$. Each distance is a fusion penalty and behaves like a grouped Lasso \citep{yuan2006grouped_lasso}. The minimization of the loss function causes some of these distances $\|\mathbf{a}_i - \mathbf{a}_j\|$ to shrink toward zero, thus interpreted as grouping objects $i$ and $j$. In related literature, this penalty is often defined using the more general $\ell_q$-norm. 
In this study, we focus on the $\ell_2$-norm since it guarantees a single partitioning of clusters for a given $\lambda$.
For fundamental work on convex clustering using the $\ell_1$-norm, see, for example, \cite{radchenko2017cvx}.

Without loss of generality, we now assume each column of $\mathbf{X}$ has a mean zero, and we propose a normalized version of \eqref{eq:clusterloss_classic}, that is, 
\begin{equation}
\label{eq:clusterloss_scaled}
    L(\mathbf{A}) = \KE \| \mathbf{X - A} \|^2 + \lambda\KP \sum_{i<j} w_{ij} \|\mathbf{a}_i - \mathbf{a}_j\|,
\end{equation}
where $\KE = (2 \| \mathbf{X} \|^2)^{-1}$ and $\KP = (\| \mathbf{X} \|  \sum_{i<j} w_{ij})^{-1}$. Normalization constants $\KE$ and $\KP$ are chosen so that the clustering of the objects in $\mathbf{X}$, given a particular value of $\lambda$, proves invariant to $n$ and $p$, as well as to the scales of $w_{ij}$ and $\mathbf{X}$. 

\begin{figure}
    \begin{minipage}{\dimexpr0.5\linewidth-1pt}
        \centering
        \includegraphics[scale=1]{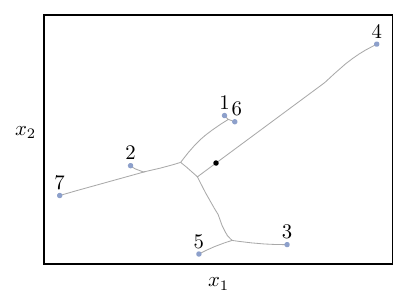}\\[-0.2cm]
        \small{(a) Clusterpath}
    \end{minipage}
    \begin{minipage}{\dimexpr0.5\linewidth-1pt}
        \centering
        \includegraphics[scale=1]{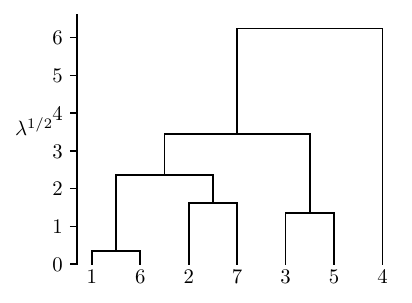}\\[-0.2cm]
        \small{(b) Clusterpath dendrogram}
    \end{minipage}
    \caption{In Panel (a): a clusterpath is computed for seven data points in $\mathbb{R}^2$ by minimizing \eqref{eq:clusterloss_scaled} for a sequence of values for $\lambda \in [0, 39]$. The objects in the $7 \times 2$ data matrix $\mathbf{X}$ are denoted by the blue dots with the paths traced by their cluster centroids in $\mathbf{A}$ appearing in grey. For $\lambda = 39$, the paths of each cluster centroid join at the central black dot. In Panel (b): the corresponding dendrogram appears using the square root of $\lambda$ as the height where objects are clustered.}
    \label{fig:cp_and_dendro}
\end{figure}

One vital property of convex clustering is its strong convexity of the loss function in $\mathbf{A}$.
In contrast to $k$-means, the minimum solution of \eqref{eq:clusterloss_scaled} with a given $\lambda\ge 0$ proves unique, and convex clustering here averts the local minima issue. Additionally, the minimization of the loss function for a sequence of values for $\lambda$ yields a series of solutions $\{ \mathbf{A}_\lambda \}$ that reflects the movement of cluster centroids through the $p$-dimensional space. These paths are commonly referred to as the \textit{clusterpath} and allow for dynamic visualizations of results \citep{allen2020carp}. Figure~\figref{fig:cp_and_dendro}{a} shows a clusterpath for a data set featuring seven objects in $\mathbb{R}^2$. In Figure~\figref{fig:cp_and_dendro}{b}, these results are converted into a dendrogram---a common visual representation in hierarchical clustering.

Convex clustering incurs several shortcomings. First, \cite{hocking2011clusterpath} have noted for particular choices of $w_{ij}$ in \eqref{eq:clusterloss_scaled} that centroids equal for some value $\lambda_1$ can reseparate for some value $\lambda_2 > \lambda_1$. This potential violation of cluster hierarchy is undesirable from an interpretative point of view and should be avoided. Second, apart from special focus on the weights, the shapes of clusters recovered by the minimization of \eqref{eq:clusterloss_scaled} tend to be spherical, which is a limiting factor in the application of the method. Third, existing algorithms scale poorly for data sets exceeding a few thousand objects. 
Prior research addressed the first two issues, demonstrating that nonconvex cluster shapes can be achieved by using $k$-nearest neighbors for nonzero weight selection \citep{chi2015ama}, and cluster separations can be mitigated by employing weights $w_{ij}$ that decrease with the distance $\| \mathbf{x}_i - \mathbf{x}_j \|$ \citep{chi2019recovering, chiquet2017fast} or cluster fusions \citep{hocking2011clusterpath, marchetti2014solution, yi2021cobrac}. The contribution of this study is twofold. First, while leveraging the established solutions for the first two issues, we tackle the third: enhancing scalability. Second, we aim to further the understanding of the effect of the weights on the clusterpath and offer two new approaches that guarantee that any number of clusters between 1 and $n$ can be attained.

Numerous algorithms have been developed to minimize the convex clustering loss function. For example, \cite{hocking2011clusterpath} introduced a subgradient descent algorithm, while \cite{chi2015ama} employed an alternating minimization algorithm (AMA) using the augmented Lagrangian. \cite{sun2018sparse_convex} developed SSNAL, using a Newton based augmented Lagrangian method. To the best of our knowledge, SSNAL is currently the fastest sequential algorithm for performing convex clustering. Another category of algorithms exploit the parallel computing capabilities of modern CPUs:
\cite{chen2015convex} proposed a parallel majorization-minimization (MM) algorithm, \cite{zhou2021scalable} introduced a parallel coordinate descent method, and \cite{fodor2022parallel} presented a parallel alternating direction method of multipliers (ADMM) algorithm. Notably, despite being a sequential method, SSNAL showed computation times comparable to those of the parallel ADMM algorithm, which used five cores for computational speed \citep{fodor2022parallel}. 

In this study, we propose a new sequential algorithm called \textit{convex clustering through majorization-minimization} (CCMM), which exhibits substantial speed improvements over SSNAL, ranging from 56 to 1,600 times faster in our experiments. An MM algorithm iteratively minimizes a simplified problem that approximates a more complex one, aptly converging toward the optimum of the complex problem \citep{hunter2004IM_tutorial}. For CCMM, we derive a novel majorization function using diagonal majorization, enabling efficient updates with a time complexity linear in $n$.

The remainder of this paper is organized as follows. Section~\ref{sec:methods} details three different approaches for choosing weights $w_{ij}$ in \eqref{eq:clusterloss_scaled}. We focus on the third (final) choice, sparse weights, to further the understanding of the effect they have on the clusterpath. In Section~\ref{sec:majorization}, we combine weight-sparsity and cluster fusions into an MM algorithm to minimize the loss function. Finally, in Sections~\ref{sec:results} and \ref{sec:conclusion}, we present results of the simulation study and conclude our paper.

\section{Hierarchy and Weights in Convex Clustering}
\label{sec:methods}
One potential pitfall of convex clustering is that increasing $\lambda$ does not always guarantee a complete cluster hierarchy. In this section, we offer a solution for this. To make convex clustering scalable for big $n$, sparsity of weights is necessary where many $w_{ij}$ are set to zero. \cite{sun2018sparse_convex} have shown that using such sparsity in weights equips their method to minimize a single instance of \eqref{eq:clusterloss_scaled} in roughly six minutes for a data set of 200,000 objects. However, under such extreme sparsity, it is likely that some groups of objects are not connected to the remaining objects via nonzero weights. Consequently, the original convex clustering problem can be reduced to separate optimization problems. It is this reducibility that renders a complete cluster hierarchy impossible. Here, we provide a formal proof of this property and craft two solutions that have minimal impact on the scalability of our algorithm. 

\subsection{Enforcing Hierarchical Clusters}
\label{sec:hierarchical_clustering}

To address cluster splits, we first note the conjecture made by \cite{hocking2011clusterpath}, which suggests that splits can be avoided if the weights $w_{ij}$ are nonincreasing in the pair-wise distances $\|\mathbf{x}_i - \mathbf{x}_j\|$. Additionally, \cite{chi2019recovering} and \cite{chiquet2017fast} provide strong theoretical support for the notion that nonincreasing weights are sufficient to prevent cluster splits.
Therefore, if the weights satisfy this condition, we can fuse the rows of $\mathbf{A}$ where $\| \mathbf{a}_i - \mathbf{a}_j \| = 0$ and store all unique centroids in $c \times p$ matrix $\mathbf{M}$. We next define $n \times c$ cluster membership matrix $\mathbf{U}$ as having elements 
\begin{equation}
    u_{ik} =
    \begin{cases}
        1 & \text{if object $i$ belongs to cluster $k$, and}\\
        0 & \text{otherwise}.
    \end{cases}\nonumber
\end{equation}
Hence, the matrix $\mathbf{A}$ can be retrieved simply via $\mathbf{A = UM}$. Instead of minimizing \eqref{eq:clusterloss_scaled} with respect to $\mathbf{A}$, we substitute $\mathbf{UM}$ for $\mathbf{A}$ and perform minimization on $\mathbf{M}$, continuing the fusion process whenever $\| \mathbf{m}_k - \mathbf{m}_l \| = 0$. 
In the context of convex biclustering, \cite{yi2021cobrac} refer to cluster fusions as \textit{compression}.

We propose to actively prevent any merged clusters from undergoing splits for larger values of the penalty parameter $\lambda$. As the algorithm proposed in Section~\ref{sec:majorization} supports warm starts as initialization, a solution $\mathbf{U}$ and $\mathbf{M}$ that minimizes \eqref{eq:clusterloss_scaled} for a specific $\lambda$  can serve as a warm start for the minimization for a subsequent value of $\lambda$. This strategy does not only ensure a hierarchy of clusters with increasing $\lambda$, it also yields a substantial gain in computational efficiency since there are fewer parameters to optimize over in $\mathbf{M}$ as compared to $\mathbf{A}$.

\subsection{Introducing Sparsity}
\label{sec:weights}
\begin{figure}
    \begin{minipage}{\dimexpr0.33\linewidth-1pt}
        \centering
        \includegraphics[scale=1]{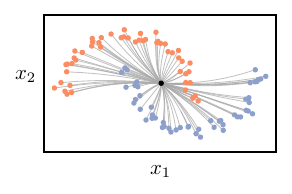}\\[-0.2cm]
        \small{(a) Unit weights}
    \end{minipage}
    \begin{minipage}{\dimexpr0.33\linewidth-1pt}
        \centering
        \includegraphics[scale=1]{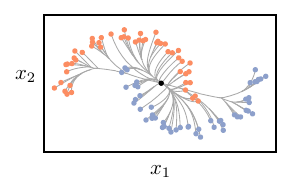}\\[-0.2cm]
        \small{(b) Gaussian weights}
    \end{minipage}
    \begin{minipage}{\dimexpr0.33\linewidth-1pt}
        \centering
        \includegraphics[scale=1]{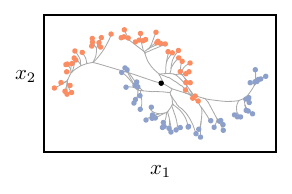}\\[-0.2cm]
        \small{(c) $k$-nn Gaussian weights}
    \end{minipage}
    \caption{Comparison of three clusterpaths computed for the same data using unit weights in the penalty for Panel (a), Gaussian weights with $\phi=4$ for Panel (b), and $k$-nn Gaussian weights with $k=10$ and $\phi=4$ for Panel (c). A black dot marks the location of the final cluster in each clusterpath.}
    \label{fig:all_weight_types}
\end{figure}

In existing literature, three general approaches for choosing the weights can be identified. The first type, used by \cite{pelckmans2005convex_first} and \cite{lindsten2011convex_also}, simply sets all $w_{ij}=1$ in a unit-weights approach. However, using convex clustering in combination with unit weights leads to quite poor clustering performance. This is illustrated by the clusterpath in Figure~\figref{fig:all_weight_types}{a}, where almost no clustering takes place before each individual path reaches the center. \cite{hocking2011clusterpath} have found that weights that decrease in the distance between $\mathbf{x}_i$ and $\mathbf{x}_j$ allow for better cluster recovery. They suggest using weights proportional to the Gaussian distribution: $w_{ij} = \exp(-\psi \| \mathbf{x}_i - \mathbf{x}_j \|^2)$ with hyperparameter $\psi$ scaling these high-dimensional Euclidean distances. 

Figure~\figref{fig:all_weight_types}{b} shows that Gaussian weights indeed outperform unit weights in terms of clustering, but fail in allowing recovery of the original two interlocking half moons. Finally, as an extension to Gaussian weights, \cite{chi2015ama} set $w_{ij}$ to zero when object $i$ is not among the $k$ nearest neighbors of $j$. This approach improves clustering performance (see Figure~\figref{fig:all_weight_types}{c}) with the added advantage of greatly reducing the computational complexity of minimization algorithms. In this paper, a scaled version of the $k$-nn Gaussian weights is used to ensure that weights remain unaffected by the scale of the data, yielding the following representation of the sparse weight matrix $\mathbf{W}$
\begin{equation}
\label{eq:sparse_symmetric_W}
w_{ij} = w_{ji} =
\begin{cases}
    \exp \left( -\phi \frac{\| \mathbf{x}_i - \mathbf{x}_j \|^2}{\mean_{i',j'}\| \mathbf{x}_{i'} - \mathbf{x}_{j'} \|^2} \right) & \text{if $j \in \mathcal{S}_i^{(k)} \lor i \in \mathcal{S}_j^{(k)}$}\\
    0 & \text{otherwise},
\end{cases}
\end{equation}
where $\mathcal{S}_i^{(k)}$ denotes the set of $k$ nearest neighbors of $i$, $\phi\geq0$ is a tuning parameter, and $\| \mathbf{x}_i - \mathbf{x}_j \|^2$ is scaled by the mean of all squared distances.

\subsection{Enforcing Irreducibility}
\label{sec:irreducibility}
Weight sparsity can improve the flexibility of convex clustering and slash computation times. However, sparsity also incurs a new challenge: the emergence of different groups of objects, for example $\mathcal{C}_k$ and $\mathcal{C}_l$ where $w_{ij} = 0 $ for all $i \in \mathcal{C}_k$ and $j \in \mathcal{C}_l$, that trigger reducibility of the convex clustering problem where a loss function can split into standalone sub-problems \citep{chi2015ama}.
Such a weight matrix we call ``disconnected" per the corresponding disconnected graph in graph theory.
The following proposition states that the minimum number of clusters is larger than one if the weight matrix is not connected.

\begin{proposition}
\label{prop:disconnected}
Let each object in $\mathbf{A}$ correspond to a node in a graph with an edge between nodes $i$ and $j$ if $w_{ij} > 0$. Let the $K$ connected components be denoted by $\mathcal{C}_k$, with $1 \leq k \leq K$. If $\mathbf{W}$ is not a connected weight matrix ($K > 1$), then there is no guarantee that all rows of $\mathbf{A}$ join into one cluster as $\lambda \rightarrow \infty$.
\end{proposition}
\begin{proof}
For disconnected $\mathbf{W}$, the loss function $L(\mathbf{A})$ equates the sum of $K$ distinct loss functions:
\begin{align*}
    L(\mathbf{A}) &= \KE \sum_{i=1}^n \| \mathbf{x}_i - \mathbf{a}_i \|^2 + \lambda\KP \sum_{i<j} w_{ij} \| \mathbf{a}_i - \mathbf{a}_j \|\\
    &= \sum_{k=1}^K \left[ \KE \sum_{i \in \mathcal{C}_k} \| \mathbf{x}_i - \mathbf{a}_i \|^2 + \lambda\KP \sum_{\substack{i,j \in \mathcal{C}_k \\i<j}} w_{ij} \| \mathbf{a}_i - \mathbf{a}_j \| \right].
\end{align*}
The minimum of this sum is obtained by independently minimizing each sub-problem. Following Proposition~2.2 in \cite{chi2015ama}, the rows of $\mathbf{A}$ can be computed as 
$\mathbf{a}_i = |\mathcal{C}_k|^{-1} \sum_{j \in \mathcal{C}_k} \mathbf{a}_j$, for all $j$ where $\{i, j\} \subseteq \mathcal{C}_k$. Therefore, it may not hold that $\mathbf{a}_i = \mathbf{a}_j$ if $\{i, j\} \not\subseteq \mathcal{C}_k$ for all $k$.
\end{proof}
Since the sparse specification of $\mathbf{W}$ in \eqref{eq:sparse_symmetric_W} does not guarantee a connected weight matrix, Proposition~\ref{prop:disconnected} indicates that it may be impossible to obtain a single cluster when performing convex clustering for large $\lambda$. Still, hierarchical clustering methods require all objects to eventually merge into a single cluster. To attain a single-cluster result from convex clustering using a sparse weight matrix, connectedness of $\mathbf{W}$ must be guaranteed.
\begin{figure}[t]
    \begin{minipage}{\dimexpr0.5\linewidth-1pt}
        \centering
        \includegraphics[scale=1]{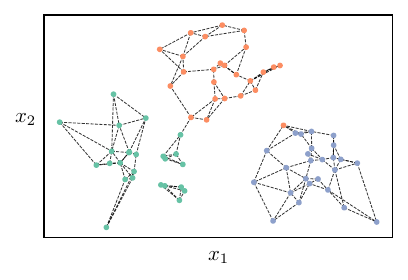}\\[-0.2cm]
        \small{(a) Disconnected weights}
    \end{minipage}
    \begin{minipage}{\dimexpr0.5\linewidth-1pt}
        \centering
        \includegraphics[scale=1]{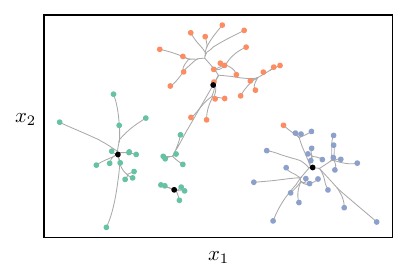}\\[-0.2cm]
        \small{(b) Disconnected clusterpath}
    \end{minipage}
    \begin{minipage}{\dimexpr0.5\linewidth-1pt}
        \centering
        \includegraphics[scale=1]{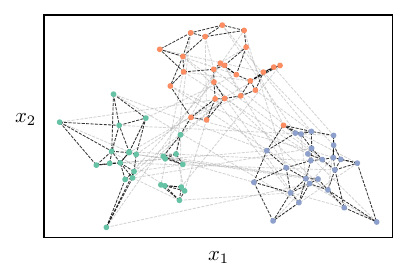}\\[-0.2cm]
        \small{(c) Symmetric circulant weights}
    \end{minipage}
    \begin{minipage}{\dimexpr0.5\linewidth-1pt}
        \centering
        \includegraphics[scale=1]{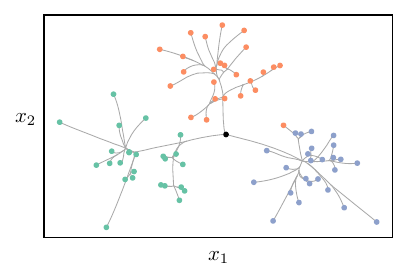}\\[-0.2cm]
        \small{(d) Symmetric circulant clusterpath}
    \end{minipage}
    \begin{minipage}{\dimexpr0.5\linewidth-1pt}
        \centering
        \includegraphics[scale=1]{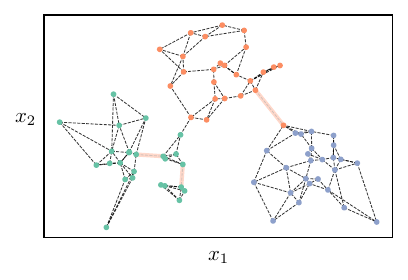}\\[-0.2cm]
        \small{(e) Minimum spanning tree weights}
    \end{minipage}
    \begin{minipage}{\dimexpr0.5\linewidth-1pt}
        \centering
        \includegraphics[scale=1]{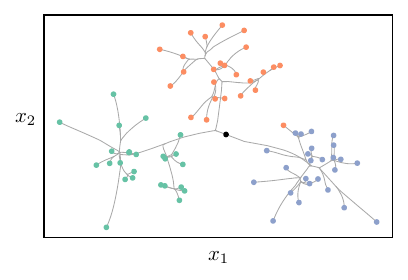}\\[-0.2cm]
        \small{(f) Minimum spanning tree clusterpath}
    \end{minipage}
    \caption{The impact of using disconnected and connected weight matrices in convex clustering. In the left column, nonzero weights are shown as dashed edges between objects in a randomly generated data set. In Panel (a): nonzero weights are computed using $k$-nn only ($k=3$). In Panel (c): connectedness via a symmetric circulant matrix, with the added nonzero weights shown in a lighter shade. In Panel (e): connectedness via a minimum spanning tree, with the additional edges highlighted for clarity. The right column (panels b, d, and f), showcases the corresponding clusterpaths that result from the use of the different weight matrices. The colors correspond to the true clusters generated by the data process, and a black dot marks the location of the final cluster(s) in each clusterpath.}
    \label{fig:sym_circ}
\end{figure}
To do so, we propose two different methods of adding nonzero weights to $\mathbf{W}$ to ensure its connectedness.

The first method avoids computationally expensive pre-processing steps.
We enlist a symmetric circulant matrix to provide a rule for adding nonzero entries to $\mathbf{W}$. A symmetric circulant is a square matrix $\mathbf{H}$ where each row is identical to the row above, but shifted one position to the right and satisfying $h_{ij}=h_{ji}$ \citep{gower1990symm_circ}. One example of the most sparse symmetric circulant \textit{not} forming the identity matrix for $n=6$ is
\begin{equation}
    \mathbf{H} = \begin{bmatrix}
    0 & 1 & 0 & 0 & 0 & 1 \\
    1 & 0 & 1 & 0 & 0 & 0 \\
    0 & 1 & 0 & 1 & 0 & 0 \\
    0 & 0 & 1 & 0 & 1 & 0 \\
    0 & 0 & 0 & 1 & 0 & 1 \\
    1 & 0 & 0 & 0 & 1 & 0 \\
    \end{bmatrix}.\nonumber
\end{equation}
As a weight matrix, $\mathbf{H}$ is connected by definition. Therefore, if nonzero weights are added to $\mathbf{W}$ according to the nonzero elements of $\mathbf{H}$, connectedness of $\mathbf{W}$ is guaranteed while adding no more than $n$ nonsparse elements. By integrating this rule into the formulation in \eqref{eq:sparse_symmetric_W}, we obtain
\begin{equation}
\label{eq:sparse_symmetric_circ_W}
w_{ij} = w_{ji} =
\begin{cases}
    \exp \left(  -\phi \frac{\| \mathbf{x}_i - \mathbf{x}_j \|^2}{\mean_{i',j'}(\| \mathbf{x}_{i'} - \mathbf{x}_{j'} \|^2)} \right) & \text{if $j \in \mathcal{S}_i^{(k)} \lor i \in \mathcal{S}_j^{(k)} \lor h_{ij}=1$},\\
    0 & \text{otherwise}.
\end{cases}
\end{equation}
Although inexpensive to construct, the nonzero elements of $\mathbf{W}$ may change depending on the order of the objects. Moreover, in the worst case scenario, it introduces $n$ additional evaluations of the distance between two rows of $\mathbf{A}$ in the penalty term in \eqref{eq:clusterloss_scaled}.

The second proposed method aims to minimize the number of added weights to arrive at a connected weight matrix. To achieve this, the method starts by identifying the $K$ connected components in $\mathbf{W}$ computed according to \eqref{eq:sparse_symmetric_W}. Then, a Gaussian weight is added for the two objects that have the smallest distance while belonging to different components. Repeating this step $K-1$ times yields a connected $\mathbf{W}$. Due to the resemblance to the procedure described in \cite{kruskal1956mst}, we refer to the resulting weight matrix as \emph{minimum spanning tree weights}. In fact, it can be easily demonstrated that the additional nonzero weights correspond to the edges derived from the construction of a minimum spanning tree for a specific graph. In this graph, the nodes are composed of the connected components, and the edges connecting these components are weighted based on the shortest distance between them.

In a graphical context, Figure~\figref{fig:sym_circ}{a} features nonzero weights based on three nearest neighbors as edges between objects in a generated data set. In Figures \figref{fig:sym_circ}{c} and \figref{fig:sym_circ}{e}, the additional nonzero weights per the symmetric circulant and the minimum spanning tree weights appear in light grey.
Figure~\ref{fig:sym_circ} also depicts the effect of different weight matrices on the clusterpath that results from minimizing \eqref{eq:clusterloss_scaled}. In Figure~\figref{fig:sym_circ}{b}, the disconnected weight matrix from \eqref{eq:sparse_symmetric_W}, using only $k$-nn, fails to produce a clusterpath leading to a complete hierarchy. In contrast, both the symmetric circulant and minimum spanning tree weights produce a weight matrix which enables a complete cluster hierarchy (see Figures~\figref{fig:sym_circ}{d} and \figref{fig:sym_circ}{f}).

Depending on the specific application, any of the three methods discussed in this paper for constructing a weight matrix can be a valid choice. If a complete cluster hierarchy is unnecessary, ensuring connectedness may not be a priority. However, if a complete hierarchy is required, users can weigh the trade-off between the number of nonzero weights added and the complexity of the preprocessing step needed to ensure connectedness.

\section{Minimization Algorithm}
\label{sec:majorization}
Cluster fusions play an important role in the CCMM algorithm by ensuring that no cluster splits occur while reducing the computational burden. Therefore, we first show in Section~\ref{sec:cluster_fusions} how fusions are enlisted to formulate the loss function in terms of the unique centroids in matrix $\mathbf{A}$.
In Section~\ref{sec:MM_update}, a surrogate function based on the reformulated loss function is derived that can be minimized analytically in linear time---making it an excellent candidate for an iterative minimization algorithm.
The convergence of an algorithm using MM to minimize a nonsmooth function like the convex clustering loss is discussed in Section~\ref{sec:convergence}.
In Section~\ref{sec:alg_implementation}, we present the complete CCMM algorithm and discuss its implementation and complexity.

\subsection{Cluster Fusions}
\label{sec:cluster_fusions}
To avoid cluster splits and to reduce the computational burden, rows of $\mathbf{A}$ that are identical can be merged. This approach tracks all unique rows of $\mathbf{A}$ and stores them in the $c \times p$ matrix $\mathbf{M}$ where $c$ denotes the number of unique rows or, equivalently, the current number of clusters.
The decision to fuse two clusters is based on whether $\mathbf{m}_k$ and $\mathbf{m}_l$ are sufficiently close given some threshold $\varepsilon_f$ \citep{hocking2011clusterpath, marchetti2014solution, yi2021cobrac}.
If $\| \mathbf{m}_k - \mathbf{m}_l \| \leq \varepsilon_f$ for some $k\neq l$,
then rows $\mathbf{m}_k$ and $\mathbf{m}_l$ in $\mathbf{M}$ are replaced by the weighted average
\begin{equation}
\label{eq:fusions_weighted_mean}
    \mathbf{m}_{new} = \frac{|C_k| \mathbf{m}_k + |C_l| \mathbf{m}_l}{|C_k| + |C_l|}.
\end{equation}
If $\mathbf{UM}$ replaces $\mathbf{A}$, we can rewrite the loss function in \eqref{eq:clusterloss_scaled} in terms of $\mathbf{M}$. As stated earlier, the advantage here is that the complexity of computing an update for $\mathbf{M}$ in an iterative algorithm no longer scales in the number of objects $n$, but in the number of clusters $c$ ranging from $n$ to 1. The new loss function becomes
\begin{align}
    L(\mathbf{M}) &= \KE \| \mathbf{X-UM} \|^2 + \lambda\KP \sum_{i<j}w_{ij} \|\mathbf{u}_i^\top \mathbf{M} - \mathbf{u}_j^\top \mathbf{M} \| \nonumber\\
    &= \KE \| \mathbf{X-UM} \|^2 + \lambda\KP \sum_{k<l}
    \mathbf{u}_{k}\T\mathbf{Wu}_{l}
    \|\mathbf{m}_k - \mathbf{m}_l \|. \label{eq:loss_function_fusions}
\end{align}
The potential increase in computational efficiency resulting from cluster fusions is substantial. Nonetheless, it is crucial to ensure that the optimum of the fused loss function closely approximates that of the unfused loss.
Since $\varepsilon_f$ is a value near zero, $\mathbf{m}_k$ and $\mathbf{m}_l$ may not equate, thus yielding a difference between $L(\mathbf{M})$ and the $L(\mathbf{M}_{new})$ that results from fusing $\mathbf{m}_k$ and $\mathbf{m}_l$. The following proposition states that as $\varepsilon_f$ approaches zero, so does $|L(\mathbf{M}) - L(\mathbf{M}_{new})|$. The proof is in Appendix~\ref{app:fusions}.

\begin{proposition}
\label{theorem:fusions}
Let the matrix $\mathbf{M}$ contain the $c$ unique rows of $\mathbf{A}$, and let $\varepsilon_f$ be the threshold used for fusing rows of $\mathbf{M}$. 
Assume that $\| \mathbf{m}_k - \mathbf{m}_l \| \leq \varepsilon_f$ for some $k\neq l$. 
Let $\mathbf{m}_{new}$ be the weighted average of $\mathbf{m}_k$ and $\mathbf{m}_l$ computed as in \eqref{eq:fusions_weighted_mean}, and let $\mathbf{M}_{new}$ be the matrix that results from setting $\mathbf{m}_k$ and $\mathbf{m}_l$ to $\mathbf{m}_{new}$. Then, for fixed $\lambda$ and $\mathbf{W}$, the absolute error $|L(\M) - L(\M_{new})|$ tends toward zero as $\varepsilon_f \rightarrow 0$.
\end{proposition}

\noindent An important consequence of Proposition~\ref{theorem:fusions} is that for sufficiently small $\varepsilon_f$ there are no order effects of fusing two rows of $\mathbf{M}$.

\subsection{Majorization-Minimization Update}
\label{sec:MM_update}
To minimize the loss function in \eqref{eq:loss_function_fusions}, we enlist MM. Here, an objective function difficult to minimize is replaced by a simpler or surrogate function with a minimum that can be computed analytically. A similar procedure was first described by \cite{weisz1937im}, but \cite{ortega1970iterative} were the first to apply it in the context of a line search. Later, it was developed as a general minimization method by
independently by \citetalias{leeuw1977IM} \citeyearpar{leeuw1977IM} and 
\cite{voss1980IM}. In machine learning literature, MM is also known as the concave-convex procedure \citep[CCCP;][]{yuille2003cccp}. Additional examples can be found in \cite{lange2000IM_optimization} and \cite{hunter2004IM_tutorial}.

MM exploits a majorization function that touches the target function at a ``supporting" point and that is never exceeded by the target function over its domain. Let $g(\mathbf{M}, \mathbf{M}_0)$ denote this majorization function with $\mathbf{M}_0$ as the supporting point. The criteria can be summarized as
\begin{equation}
\label{eq:maj_conditions}
    g(\mathbf{M}, \mathbf{M}_0) \geq L(\mathbf{M}) \quad \text{and} \quad g(\mathbf{M}_0, \mathbf{M}_0) = L(\mathbf{M}_0).
\end{equation}
The key is to choose a function for $g(\mathbf{M}, \mathbf{M}_0)$ that is quadratic in $\mathbf{M}$ where its minimum can be computed analytically for use in the next iteration as the new supporting point. 
Under some relatively weak conditions, this iterative algorithm converges to the minimum of the target function (see Section~\ref{sec:convergence}).

The first step in defining $g(\mathbf{M}, \mathbf{M}_0)$ is to majorize the $\ell_2$-norm of the differences between rows of $\mathbf{M}$. 
If $a,b\geq 0$, then $(ab)^{1/2}\leq (a+b)/2$ since $0\leq (a^{1/2}-b^{1/2})^2$.
Substituting $a=\| \bm \theta \|^2$ and $b= \| \bm \theta_0 \|^2$ yields 
\begin{equation}
\label{eq:major_norm}
    \|\bm{\theta}\| \leq \frac{1}{2} \frac{\|\bm{\theta}\|^2}{\|\bm{\theta}_0\|} + \frac{1}{2}\|\bm{\theta}_0\|,
\end{equation}
for $\|\bm{\theta}_0\| > 0$. Applying this inequality to $ \| \mathbf{m}_k - \mathbf{m}_l \| $, using $\| \mathbf{m}_k - \mathbf{m}_l \|^2 = (\mathbf{m}_k - \mathbf{m}_l)\T(\mathbf{m}_k - \mathbf{m}_l) = \tr (\mathbf{m}_k - \mathbf{m}_l)(\mathbf{m}_k - \mathbf{m}_l)\T = \tr \mathbf{M}\T\mathbf{E}_{kl}\mathbf{M}$ subject to $\mathbf{E}_{kl} = (\mathbf{e}_k-\mathbf{e}_l)(\mathbf{e}_k-\mathbf{e}_l)\T$ with $\mathbf{e}_k$ being the $k$\textsuperscript{th} column of the $c \times c$ identity matrix, finally yields
\begin{align}
    L(\mathbf{M}) \opl{\leq} \KE\| \mathbf{X-UM} \|^2  \nonumber+ \frac{\lambda \KP}{2} \sum_{k<l} %(\UWU)_{kl} 
    \mathbf{u}_k^\top\mathbf{Wu}_l
    \left(\frac{\|\mathbf{m}_k - \mathbf{m}_l \|^2}{\|\mathbf{m}_{0,k} - \mathbf{m}_{0,l} \|} + \|\mathbf{m}_{0,k} - \mathbf{m}_{0,l} \| \right) \nonumber\\
    \opl{=} \KE \| \mathbf{X-UM} \|^2 + \frac{\lambda\KP}{2} \tr \mathbf{M}\T\mathbf{C}_0\mathbf{M} + c(\mathbf{M}_0),\label{eq:mm_function_quadratic}
\end{align}
where
\begin{equation*}
    \mathbf{C}_0 = \sum_{k<l} \frac{
    %(\UWU)_{kl} 
    \mathbf{u}_k^\top\mathbf{Wu}_l
    } {\| \mathbf{m}_{0,k} - \mathbf{m}_{0,l} \|} \mathbf{E}_{kl},
\end{equation*}
with $c(\mathbf{M}_0)$ the collection of constant terms not depending on $\mathbf{M}$. We should note that using cluster fusions in the algorithm prevents a division by zero in the computation of $\mathbf{C}_0$.

At this point, the function in \eqref{eq:mm_function_quadratic} is quadratic in $\mathbf{M}$ and is continuously differentiable, aptly suiting it for use as $g(\mathbf{M}, \mathbf{M}_0)$ in the MM algorithm. Here, the update for $\mathbf{M}$ satisfies
\begin{align}
    \frac{\partial g(\mathbf{M}, \mathbf{M}_0)}{\partial \mathbf{M}} = 2 \KE \mathbf{U}\T\mathbf{UM} - 2 \KE \mathbf{U}\T\mathbf{X} + \lambda\KP \mathbf{C}_0\mathbf{M} &= \mathbf{O}\nonumber\\
    \left( \mathbf{U}\T\mathbf{U} + \frac{\lambda\KP}{2 \KE } \mathbf{C}_0 \right)\mathbf{M} &= \mathbf{U}\T\mathbf{X}, \label{eq:fusions_update}
\end{align}
where $\mathbf{O}$ is a matrix of zeroes. However, solving the system of equations in \eqref{eq:fusions_update} incurs a complexity of \bigO{c^3}---a costly operation when $c$ is large. As $\mathbf{U}\T\mathbf{U}$ is a diagonal matrix, the issue lies with the matrix $\mathbf{C}_0$. We thus take another step in the majorization process by eliminating $\tr \mathbf{M}\T\mathbf{C}_0\mathbf{M}$ from \eqref{eq:mm_function_quadratic}. \cite{groenen1999diag_major} treated a similar problem by carrying out a majorization step using upper the bounds of the eigenvalues of $\mathbf{C}_0$. If $\mathbf{D}_0$ is a diagonal matrix that features these upper bounds as its elements, then $\mathbf{D}_0 - \mathbf{C}_0$ is positive semi-definite. Consequently, the following inequality can be derived
\begin{align*}
    0 &\leq \tr [ (\mathbf{M}-\mathbf{M}_0)\T(\mathbf{D}_0 - \mathbf{C}_0)(\mathbf{M}-\mathbf{M}_0) ]\\
    \tr \mathbf{M}\T\mathbf{C}_0\mathbf{M} &\leq \tr \mathbf{M}\T\mathbf{D}_0\mathbf{M} - 2 \tr \mathbf{M}\T(\mathbf{D}_0-\mathbf{C}_0)\mathbf{M}_0 + \tr \mathbf{M}_0\T(\mathbf{D}_0-\mathbf{C}_0)\mathbf{M}_0.
\end{align*}
To obtain $\mathbf{D}_0$, Gershgorin's Circle Theorem \citep{bronson1989gershgorin} can be used where eigenvalues of $\mathbf{C}_0$ fall within $[ c_{kk} - \sum_{l \neq k} |c_{kl}|, c_{kk} + \sum_{l \neq k} |c_{kl}| ]$. Due to the structure of $\mathbf{C}_0$, the upper bound $c_{kk} + \sum_{l \neq k} |c_{kl}| = 2 c_{kk}$, which means that $\mathbf{D}_0$ can be computed as twice the diagonal of $\mathbf{C}_0$. Substituting the inequality for $\tr \mathbf{M}\T\mathbf{C}_0\mathbf{M}$ into the majorized loss function in \eqref{eq:mm_function_quadratic} yields
\begin{align}
    L(\mathbf{M}) \opl{\leq} \KE \| \mathbf{X-UM} \|^2 + \frac{\lambda\KP}{2} \tr \mathbf{M}\T\mathbf{D}_0\mathbf{M} - \lambda\KP \tr \mathbf{M}\T(\mathbf{D}_0-\mathbf{C}_0)\mathbf{M}_0 + c(\mathbf{M}_0) \nonumber\\
    \opl{=} g(\mathbf{M}, \mathbf{M}_0). \label{eq:mm_function}
\end{align}
Then, the matrix $\mathbf{M}$ that minimizes $g(\mathbf{M}, \mathbf{M}_0)$ can be computed as
\begin{align}
    \frac{\partial g(\mathbf{M}, \mathbf{M}_0)}{\partial \mathbf{M}} \opl{=}  2 \KE \mathbf{U}\T \mathbf{UM} - 2 \KE \mathbf{U}\T\mathbf{X} + \lambda\KP \bigl( \mathbf{D}_0 \mathbf{M} -  (\mathbf{D}_0 - \mathbf{C}_0)\mathbf{M}_0 \bigr) = \mathbf{O}\nonumber\\
    \mathbf{M} \opl{=} \left(\mathbf{U}\T\mathbf{U} + \frac{\lambda\KP}{2 \KE} \mathbf{D}_0 \right)^{-1} \left( \mathbf{U}\T\mathbf{X} + \frac{\lambda\KP}{2 \KE} (\mathbf{D}_0 - \mathbf{C}_0) \mathbf{M}_0 \right). \label{eq:diag_fusions_convex_update}
\end{align}
Note that the first term is a diagonal matrix with elements
\begin{equation*}
    \left(\mathbf{U}\T\mathbf{U} + \frac{\lambda\KP}{2 \KE}  \mathbf{D}_0 \right)_{kk}^{-1} = \left(|\mathcal{C}_k| + \frac{\lambda\KP}{2 \KE}  d_{0, kk} \right)^{-1} 
\end{equation*}
that permits straightforward updates. In Section~\ref{sec:alg_implementation}, we present the complete algorithm that uses this update to perform convex clustering.

\subsection{Convergence}
\label{sec:convergence}
In this section, we demonstrate the convergence of an MM algorithm when applied to the loss function in \eqref{eq:loss_function_fusions} using the update strategy outlined in \eqref{eq:diag_fusions_convex_update}. Our objective is to demonstrate the algorithm's convergence to the global minimum of the target function. Furthermore, we discuss both the theoretical order of convergence and a method to empirically determine it.

To establish the convergence to the global minimum, we rely on several key properties and concepts. First, we consider the majorization function $g(\mathbf{M}, \mathbf{M}_0)$ in \eqref{eq:mm_function}, which is constructed to satisfy \eqref{eq:maj_conditions}. Hence, it has the nonincrement property
\begin{equation*}
    L(\mathbf{M}^{(t+1)}) \leq g(\mathbf{M}^{(t+1)}, \mathbf{M}_0) \leq g(\mathbf{M}^{(t)}, \mathbf{M}_0) = L(\mathbf{M}^{(t)}),
\end{equation*}
where $\mathbf{M}^{(t)}$ represents the $t$\textsuperscript{th} iterate and which states that $\mathbf{M}^{(t+1)} = \argmin_\mathbf{M} g(\mathbf{M}, \mathbf{M}_0)$ does not increase the target loss function. Additionally, the strong convexity of $g(\mathbf{M}, \mathbf{M}_0)$ implies that the sequence $\{ L(\mathbf{M}^{(t)}) \}$ exhibits the sufficient descent property \citep{xu2016relaxed}, which is stronger than the nonincrement property.
Moreover, \cite{xu2016relaxed} use the sufficient descent property to establish the convergence of $\{ \mathbf{M}^{(t)} \}$ to a stationary point for a range of nonsmooth functions. Because $L(\mathbf{M})$ is strictly convex and coercive, the stationary point $\mathbf{M}^* = \lim_{t \rightarrow \infty} \mathbf{M}^{(t)}$ is the global minimum (\citealp{bolte2016majorization}; \citealp[Chapter 12]{lange2013optimization}; \citealp{razaviyayn2013unified}).

While often referred to as an algorithm, MM should be regarded as a fundamental principle that can be used in optimization algorithms. Notably, the order of convergence can vary when different majorization functions are used to optimize the same target function. Typically, a well-selected majorization function leads to asymptotic linear convergence, as demonstrated in \citetalias{leeuw1988convergence} \citeyearpar{leeuw1988convergence}, \citetalias{leeuw1994mm_convergence} \citeyearpar{leeuw1994mm_convergence}, \cite{mairal2013stochastic}, and \cite{mairal2015incremental}.
However, it has been shown that during the first few iterations the asymptotic linear rate does not apply, and convergence may be faster \citep{havel1991mm_convergence}.

As part of our research, we empirically estimate the order of convergence of the CCMM algorithm. Following \cite{cordero2007variants}, the order of convergence of a sequence can be estimated by
\begin{equation}
\label{eq:orderofconvergence}
    \hat{q} \approx \frac{\log \bigl( \| \mathbf{A}^{(t+1)} - \mathbf{A}^{(t)} \| / \| \mathbf{A}^{(t)} - \mathbf{A}^{(t-1)} \| \bigr) }{\log \bigl( \| \mathbf{A}^{(t)} - \mathbf{A}^{(t-1)} \| / \| \mathbf{A}^{(t-1)}) - \mathbf{A}^{(t-2)} \| \bigr) },
\end{equation}
where $\mathbf{A}^{(t)}$ is calculated using $\mathbf{M}^{(t)}$.

\subsection{Implementation}
\label{sec:alg_implementation}
In Algorithm~\ref{alg:ccmm}, we present the CCMM algorithm used to minimize \eqref{eq:loss_function_fusions}. The algorithm is divided into six steps, each performing a straightforward task. 
\AlgStep{alg:ccmm}{1} initializes several variables used in the minimization. Notable are $\widetilde{\mathbf{X}}$ and $\widetilde{\mathbf{W}}$ used to track $\mathbf{U}\T\mathbf{X}$ and $\UWU$ to avoid unnecessary computations. The main loop of the algorithm starts by computing the update per \eqref{eq:diag_fusions_convex_update} in \AlgStep{alg:ccmm}{2}. 
To reduce the number of iterations required to reach convergence, we make use of a relaxed update (\citetalias{deleeuw1980step_doubling}, \citeyear{deleeuw1980step_doubling}), which we call ``step doubling". When step doubling, the update for $\mathbf{M}$ is computed as $\mathbf{M}^{(t+1)} = 2\mathbf{M}^{(t+1)} - \mathbf{M}_0$, where $\mathbf{M}^{(t+1)}$ is the result from \AlgStep{alg:ccmm}{2} and $\mathbf{M}_0$ the supporting point. In practice, this type of update is applied after a number of burn-in iterations $b$ to avoid ``stepping over" the minimum. This accelerated update is computed in \AlgStep{alg:ccmm}{3}, and in our experiments we typically set $b$ to 25.

In \AlgStep{alg:ccmm}{4}, rows in $\mathbf{M}^{(t+1)}$ are fused. As \cite{hocking2011clusterpath} have noted, the ideal approach for determining fusions would be to evaluate the change in loss function per centroid pair. However, this may not be viable when it incurs a large number of additional computations. Therefore, the authors suggest using a threshold that triggers a cluster fusion when $\mathbf{M}^{(t+1)}$ for which $\bigl\|\mathbf{m}^{(t+1)}_k - \mathbf{m}^{(t+1)}_l \bigr\| \leq \varepsilon_f$. We propose a formulation of $\varepsilon_f$ that reflects the scale of the data while also securing a threshold not too small in cases of very similar (duplicate) objects. Here, $\varepsilon_f$ computes as
\begin{equation}
\label{eq:fuse_thresh}
    \varepsilon_{f} = \tau \median_{i,j} \left( \|\mathbf{x}_i-\mathbf{x}_j\| \right),
\end{equation}
where $\tau$ denotes a specified fraction of the median of Euclidean distances between objects. For very large data sets, the median of medians may estimate this value. In our experiments, we typically set $\tau$ to $10^{-3}$.
\begin{Ualgorithm}
\caption{Convex clustering through majorization-minimization (CCMM).
Key parts of the algorithm are Step 2 where the update derived in \eqref{eq:diag_fusions_convex_update} is computed, and Step 4 where cluster fusions are performed. Step 3 is executed only after a number of burn-in iterations to accelerate convergence. Remaining steps perform initializations and bookkeeping.}
\label{alg:ccmm}
\KwIn{$\mathbf{X}$, $\mathbf{W}$, $\lambda$, $\varepsilon_c$, $\varepsilon_f$, $b$}
\KwOut{$\mathbf{A}$}
\BlankLineA
$\mathbf{M}_0 := \mathbf{X}$ \Comment*[r]{Step 1}
$\mathbf{U} := \mathbf{I}_n$\;
$ L_1 := L(\mathbf{M}_0)$\;
$ L_0 := (1+2 \varepsilon_c)L_1 $ \\
$\widetilde{\mathbf{X}} := \mathbf{X}$ \;
$\widetilde{\mathbf{W}} := \mathbf{W}$ \;
$\KE := \bigl({2 \| \mathbf{X} \|^2}\bigr)^{-1}$ \;
$\KP := \bigl({\| \mathbf{X} \|  \sum_{i<j} w_{ij}}\bigr)^{-1}$ \;
$\gamma := \lambda\KP/(2\KE)\;
$\;
$t := 0$\;
\BlankLineA
\While{$ ( L_0 - L_1 )/L_1 > \varepsilon_c $}{
    $\mathbf{C}_0 :=
    \sum_{k=2}^{c} \sum_{l=1}^{k-1} (\widetilde w_{kl}/\| \mathbf{m}_{0,k} - \mathbf{m}_{0,l} \|) \mathbf{E}_{kl}
    $ \Comment*[r]{Step 2}
    \vspace{1mm}
    $\mathbf{D}_0 := \diag(\mathbf{C}_0)$\;
    \vspace{0.1mm}
    $\mathbf{M}^{(t+1)} := \bigl(\mathbf{U}\T\mathbf{U} + \gamma \mathbf{D}_0 \bigr)^{-1} \bigl( \widetilde{\mathbf{X}} + \gamma (\mathbf{D}_0 - \mathbf{C}_0) \mathbf{M}_0 \bigr) $\;
    \BlankLineA
    \If(\Comment*[f]{Step 3}){$t>b$}{
        $\mathbf{M}^{(t+1)} := 2\mathbf{M}^{(t+1)} - \mathbf{M}_0$\;
    }
    \BlankLineA
    \While(\Comment*[f]{Step 4}){$\exists (k,l): \bigl\| \mathbf{m}^{(t+1)}_k - \mathbf{m}^{(t+1)}_l \bigr\| \leq \varepsilon_f$}{
        Fuse rows of $\mathbf{M}^{(t+1)}$ and obtain $\mathbf{U}_{new}$ \;
        $\widetilde{\mathbf{X}} := \mathbf{U}_{new}\T \widetilde{\mathbf{X}}$ \;
        $\widetilde{\mathbf{W}} := \mathbf{U}_{new}\T \widetilde{\mathbf{W}}\mathbf{U}_{new}$ \;
        $\mathbf{U} := \mathbf{U}\mathbf{U}_{new}$\;
    }
    \BlankLineA
    $ \mathbf{M}_0 := \mathbf{M}^{(t+1)} $ \Comment*[r]{Step 5}
    $ L_0 := L_1 $\;
    $ L_1 := L(\mathbf{M}_0)$\;
    $t := t + 1 $\;
}
\BlankLineA
$\mathbf{A} := \mathbf{U} \mathbf{M}_0$\ \Comment*[r]{Step 6}
\end{Ualgorithm}
The fusion process outputs the $c_{old} \times c_{new}$ membership matrix $\mathbf{U}_{new}$ that maps the old clusters to the new ones. This matrix can be used to update $\widetilde{\mathbf{X}}$ and $\widetilde{\mathbf{W}}$ as $\mathbf{U}_{new}\T \widetilde{\mathbf{X}}$ and $\mathbf{U}_{new}\T \widetilde{\mathbf{W}} \mathbf{U}_{new}$, respectively. Additionally, \AlgStep{alg:ccmm}{4} repeats if there are rows in the updated matrix $\mathbf{M}^{(t+1)}$ for which $\bigl\|\mathbf{m}^{(t+1)}_k - \mathbf{m}^{(t+1)}_l \bigr\| \leq \varepsilon_f$.

In \AlgStep{alg:ccmm}{5}, the loss function value is updated. When the relative decrease in the loss exceeds some value $\varepsilon_c$, the algorithm reverts to \AlgStep{alg:ccmm}{2}. At convergence, the resulting matrix $\mathbf{A}$ is computed and returned to the user. For the results presented in this paper, we set the value of $\varepsilon_c$ to $10^{-6}$.
Finally, when computing a clusterpath for multiple values of $\lambda$, warm starts can be used by setting $\mathbf{M}_0$ to the solution of the minimization for the previous value of $\lambda$ in \AlgStep{alg:ccmm}{1}.

To analyze algorithmic complexity, we focus on two of the six steps. First, in \AlgStep{alg:ccmm}{2}, the update derived in \eqref{eq:diag_fusions_convex_update} is computed. A key feature of this update is that it can be done in a single loop over the nonzero elements of the $c\times c$ matrix $\widetilde{\mathbf{W}}$. Hence, this step has a complexity of \bigO{npk} where $n$ is an upper bound on $c$. Second, in \AlgStep{alg:ccmm}{4}, the rows of $\mathbf{M}^{(t+1)}$ fuse, and several matrices update. Using the nonzero elements of $\widetilde{\mathbf{W}}$, a single pass through \AlgStep{alg:ccmm}{4} also has complexity \bigO{npk}. However, since \AlgStep{alg:ccmm}{4} may recur, it can be shown in the absolute worst case that the number of passes is $n$, and worst-case complexity is \bigO{n^2pk}.

The CCMM algorithm was coded in C\texttt{++} while making extensive use of the linear algebra library \textit{Eigen} \citep{eigen}. We used \textit{Rcpp} \citep{eddel2011rcpp, eddel2013rcpp, eddel2017rcpp} to interface the program with the programming language R and \textit{pybind11} \citep{pybind11} to interface it with Python.\footnote{Software packages that implement CCMM for both R and Python are available at \url{https://github.com/djwtouw/CCMMR} and \url{https://github.com/djwtouw/CCMMPy}.}
A key factor in scalability is sparsity in the weights as defined in \eqref{eq:sparse_symmetric_circ_W}. To efficiently find the $k$ nearest neighbors of each object, we enlisted a $k$-d tree \citep{bentley1975kdtree} implemented by \cite{arya2019knn} and \cite{scikit-learn}. 

\FloatBarrier
\section{Numerical Experiments}
\label{sec:results}
In this section, we present results of several numerical experiments. The main contribution of our research is an algorithm that performs convex clustering more efficiently than prior published algorithms. Therefore, we first present a comparison between CCMM versus its main competitors. All computational results in this paper were obtained using a desktop featuring an AMD Ryzen 5 2600X processor at 3.6 GHz using 32 GB of memory running Ubuntu version 22.04.1.

\subsection{Simulated Data}
\label{sec:twohalfmooncomparison}
To be competitive, CCMM should minimize the convex clustering loss function faster than state-of-the-art convex clustering algorithms including, to our knowledge, today's fastest approach: SSNAL \citep{sun2018sparse_convex}. Beyond comparing CCMM to SSNAL, we consider AMA \citep{chi2015ama}, which is available as an R-package.\footnote{The source code for AMA can be found at \url{https://cran.r-project.org/src/contrib/Archive/cvxclustr/} and the source code for SSNAL at \url{https://blog.nus.edu.sg/mattohkc/softwares/convexclustering/}. Additionally, the code to generate the results presented in this section is available at \url{https://github.com/djwtouw/CCMM-paper}.}

For our contest of three algorithms, we chose an approach similar to that described in \cite{sun2018sparse_convex}. Data sets ranging from 1,000 to 5,000 objects were generated per the two interlocking half-moons data generating process (see Figure~\ref{fig:all_weight_types}). For each value of $n$, we used ten different realizations of the data.
Since AMA and SSNAL typically minimize the unscaled loss in \eqref{eq:clusterloss_classic}, we adjusted the scaling constants in CCMM so that each algorithm minimized the same problem. Additionally, as the software package of SSNAL does not allow for custom (connected) weight matrices, we used the weights from \eqref{eq:sparse_symmetric_W} without the scaling component, opting to set $\phi=2$ and $k=15$.
The final input common to the three algorithms was the sequence for $\lambda$ that describes the clusterpath, set to $\{ 0.0, 0.2, \ldots, 109.8, 110.0\}$.

For the CCMM algorithm, we set $\varepsilon_c$ to $10^{-6}$ and $b$ to 25, computing $\varepsilon_f$ according to \eqref{eq:fuse_thresh} with $\tau=10^{-3}$. For AMA, the default tolerance for convergence of $10^{-3}$ was used along with the default step size parameter $\nu$ (which depends on $\mathbf{W}$ and $n$). For parameters in the SSNAL algorithm, we applied default values provided by the authors. The tolerance for terminating the algorithm was set to $10^{-6}$. For warm starts, 50 ADMM iterations were used for the first value of $\lambda$ and 20 iterations for subsequent values.

\begin{figure}
    \begin{minipage}{\dimexpr0.5\linewidth-1pt}
        \centering
        \includegraphics[scale=1]{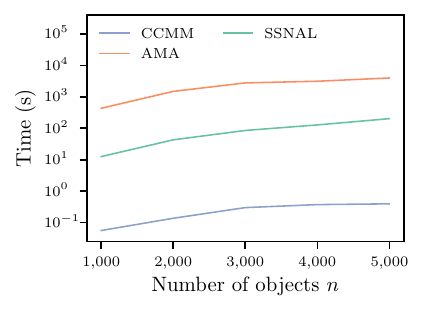}\\[-0.2cm]
        \small{(a) Average elapsed time per clusterpath}
    \end{minipage}
    \begin{minipage}{\dimexpr0.5\linewidth-1pt}
        \centering
        \includegraphics[scale=1]{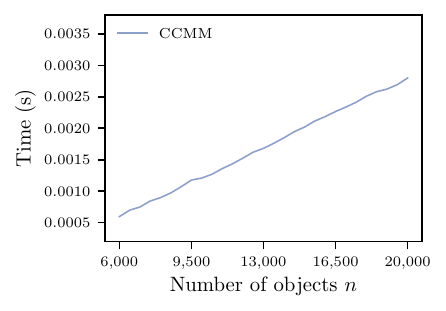}\\[-0.2cm]
        \small{(b) Average elapsed time per CCMM iteration}
    \end{minipage}
    \caption{In Panel (a): the mean computation times for performing convex clustering on the two interlocking half moon data for the number of objects ranging from 1,000 to 5,000. In Panel (b): the mean computation time per iteration of the CCMM algorithm for the two interlocking half moons ranging from 6,000 to 20,000 objects.}
    \label{fig:cpu_time}
\end{figure}

In Figure~\figref{fig:cpu_time}{a}, we present mean elapsed times obtained experimentally. Results show that for $n=$ 1,000, CCMM was roughly 7,678 times faster than AMA and 224 times faster than SSNAL. For $n=$ 5,000, speed factors rose to 10,023 and 510 times, respectively. This convincingly shows that CCMM is the fastest algorithm to perform convex clustering. 

To investigate the scalability of the CCMM algorithm, we continued the analysis of the two half-moon data sets for $n$ ranging from 6,000 to 20,000 using CCMM only. To analyze both the update of $\mathbf{M}$ and the process of cluster fusions, clusterpaths were computed until the number of clusters was roughly 90\% of $n$. Figure~\figref{fig:cpu_time}{b} shows the average elapsed time per CCMM iteration for each of the data sets, which was repeated for five different realizations of the data for each value of $n$.
Even though we derived in Section~\ref{sec:alg_implementation} each iteration's theoretical worst-case complexity to be \bigO{n^2pk}, we observe a trend linear in $n$.

\begin{figure}
    \begin{minipage}{\dimexpr0.5\linewidth-1pt}
        \centering
        \includegraphics[scale=1]{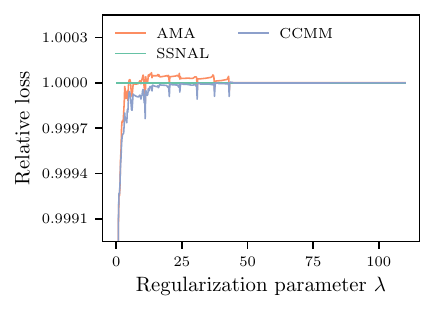}\\[-0.2cm]
        \small{(a) Relative losses for $n=$ 1,000}
    \end{minipage}
    \begin{minipage}{\dimexpr0.5\linewidth-1pt}
        \centering
        \includegraphics[scale=1]{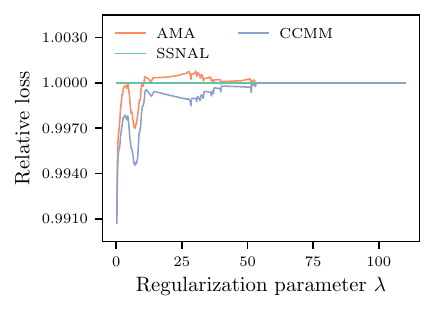}\\[-0.2cm]
        \small{(b) Relative losses for $n=$ 5,000}
    \end{minipage}
    \caption{In Panel (a): the mean value of the loss function obtained by CCMM, SSNAL, and AMA relative to the value obtained by SSNAL, using default convergence criteria. Computed for ten realizations of the two interlocking half moon data with $n=$ 1,000. In Panel (b): the same as in Panel (a), but for $n = $ 5,000.}
    \label{fig:accuracy2}
\end{figure}

To be a viable alternative to SSNAL, a new algorithm should not only minimize the loss function in less time, but also yield a similar solution. To assess the solution quality of AMA, SSNAL and CCMM, we compared values of the loss function that each algorithm obtained for data sets comprising 1,000 and 5,000 objects. In Figure~\ref{fig:accuracy2}, we use the SSNAL output as a benchmark for reporting the relative values obtained by AMA and CCMM. 
Being the slowest, AMA also most often incurred the largest value for the loss function. As Figure~\figref{fig:accuracy2}{b} shows, the relative quality of the minimum obtained by CCMM was affected positively by the sample size. On average, the minima attained by CCMM were at best 1.1650\% lower and at worst 0.0008\%  higher than those attained by SSNAL for both $n=$ 1,000 and $n=$ 5,000.

\subsection{Empirical Data}
\label{sec:res_real_data}
Besides simulated data, we used several empirical data sets to compare the three algorithms: the banknotes authentication, musk and MAGIC gamma telescope data sets \citep{dua2019uci}, and the MNIST data set \citep{deng2012mnist}. In Table~\ref{tab:real_data}, the first two columns feature the number of objects ($n$) and variables ($d$) for each data set. To attain reasonable computation times, we used UMAP \citep{mcinnes2018umap} to reduce dimensionality to the number of categories in the data ($p$). Next, the weight matrices were computed using $k=50$ and $\phi=2$. To determine the sequence for $\lambda$, we first obtained the smallest value where the number of clusters stopped decreasing. Then, the sequence was chosen as 200 values along a cubic function from $0$ to $\lambda_{max}$. Notably, omitting a symmetric circulant matrix to ensure a connected weight matrix meant that \textit{none} of the clusterpaths terminated in a single cluster. The fourth column lists the minimum number of clusters $c_{min}$.

\begin{table}
\centering
\caption{Clusterpaths for real data sets computed by three different algorithms. Dimension reduction was applied to each $n \times d$ data set using UMAP where the target number of dimensions ($p$) was the true number of clusters. Each clusterpath was computed for 200 values of $\lambda$ that terminated at the minimum number of clusters attainable ($c_{min}$). This value exceeds one since the software package for SSNAL forbids the use of custom weight matrices. The three rightmost columns list the number of seconds required by each algorithm to compute the final clusterpath. A hyphen means that the algorithm required more than the available 32 GB system memory to complete the analysis.}
\label{tab:real_data}
\begin{tabular}{l c rrrr c rrr}
\toprule
&&&&&&& \multicolumn{3}{c}{Elapsed time (s)}\\
\cmidrule{8-10}
	        && $n$	    & $d$	 & $p$	 & $c_{min}$	 && AMA	 & SSNAL	 & CCMM	 \\ \midrule 
Banknotes	&& 1,372 	& 4 	 & 2 	 & 7 	 && 380.03 	& 55.58 	& 0.20 	 \\ 
Musk	    && 6,598 	& 166 	 & 2 	 & 27 	 && - 	    & 177.02 	& 0.11 	 \\ 
Gamma	    && 19,020 	& 10 	 & 2 	 & 2 	 && - 	    & 7,266.07 	& 14.78 	 \\ 
MNIST	    && 60,000 	& 784 	 & 10 	 & 6 	 && - 	    & - 	    & 184.97 	 \\ 
\bottomrule
\end{tabular}
\end{table}

The number of seconds passed for each of the algorithms while computing the clusterpaths is reported in the three rightmost columns of Table~\ref{tab:real_data}. A missing result signals insufficient memory (32 GB system) for that algorithm. Results show that the speedup of CCMM over the other algorithms is dependent on the data analyzed. CCMM is about 1,900 times faster than AMA for the banknotes data, which differs significantly from the results for the generated data sets. The same holds for the speedup with respect to SSNAl, which ranges from roughly 280 to 1,600.

\subsection{Comparison without Cluster Fusions}
In Sections \ref{sec:twohalfmooncomparison} and \ref{sec:res_real_data}, we saw that CCMM outperforms AMA and SSNAL in the computation of clusterpaths. Cluster fusions play an important role in the observed speedups, as they reduce the size of the problem substantially as the number of clusters continues to decrease. However, cluster fusions are not unique to CCMM and, in theory, could be implemented for other algorithms as well. Therefore, in this section, we adopt a deliberate approach to avoid cluster fusions in CCMM to isolate the speedup due to our efficient MM update.

The setup of the experiment was as follows. Avoiding early fusions was done by generating objects in two-dimensional space following a grid pattern to ensure that the distance between all objects was sufficiently large. To introduce some level of randomness and prevent ties during the computation of the $k$ nearest neighbors for each object, a small amount of noise was added to the coordinates of each point. 
Data sets ranging from 500 to 20,000 objects were generated. For each value of $n$, we used ten different realizations of the data. 

The three algorithms shared the following common inputs. The parameters determining the weight matrix were $\phi=0.5$ and $k=15$. To set the sequence for $\lambda$, we first identified the value of $\lambda$ for which the first cluster fusion occurred, denoted by $\lambda_\text{max}$. Then, the sequence for $\lambda$ that describes the clusterpath was set to five evenly spaced values ranging from $0$ to $\lambda_\text{max}$. Thus, ensuring that CCMM would not perform cluster fusions during the minimizations of the loss function. For the algorithm-specific parameters we used those described in Section~\ref{sec:twohalfmooncomparison}.

Since all three algorithms in this comparison use different stopping criteria, we split the experiment in two parts. First CCMM was compared to AMA, and then to SSNAL. The stopping criterion for CCMM was determined by the value of the loss function that was attained by the opposing algorithm as 
\begin{equation*}
    L_\text{CCMM} - L_\text{Opposing} \leq 10^{-6} L_\text{Opposing},
\end{equation*}
which is the approach used by \cite{sun2018sparse_convex} to ensure a fair comparison between SSNAL and AMA.

\begin{figure}
    \begin{minipage}{\dimexpr0.5\linewidth-1pt}
        \centering
        \includegraphics[scale=1]{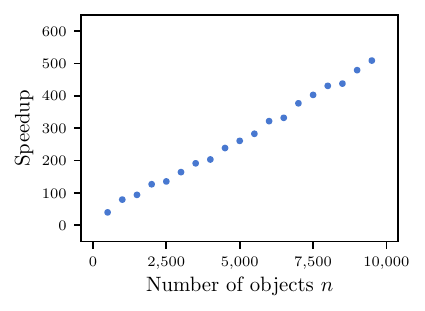}\\[-0.2cm]
        \small{(a) Elapsed times AMA/CCMM}
    \end{minipage}
    \begin{minipage}{\dimexpr0.5\linewidth-1pt}
        \centering
        \includegraphics[scale=1]{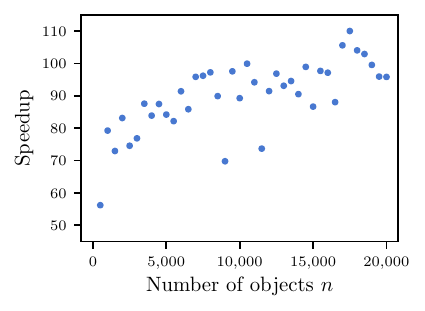}\\[-0.2cm]
        \small{(b) Elapsed times SSNAL/CCMM}
    \end{minipage}
    \caption{
        Comparative performance analyses, showcasing the relative performance gains of CCMM without fusions over both AMA and SSNAL. To compare the speed and scalability of the minimization algorithms, only the parts of the clusterpaths were used in which no clusters were formed. This approach avoided the cluster fusions of CCMM and ensured that each algorithm had to compute updates for an $n \times p$ matrix.
        Additionally, in both comparisons, the stopping criterion for CCMM was determined by the value of the loss function that was attained by the opposing algorithm on the same data.
        The length of each clusterpath was five values of $\lambda$ and the number of repetitions was ten for each value of $n$.
    }
    \label{fig:cpu_time_ratios}
\end{figure}

In Figure~\figref{fig:cpu_time_ratios}{a}, we present the ratio of the elapsed computation times for AMA and CCMM. The number of objects does not go beyond 9,500, as the available system memory (32 GB) was insufficient for AMA. The graph clearly displays an upward linear trend in the speedup of CCMM over AMA, reaching its maximum of roughly 500 for the largest data sets. The results of the comparison between SSNAL and CCMM are shown in Figure~\figref{fig:cpu_time_ratios}{b}. On average, the speedup of CCMM over SSNAL ranged from 79 for the data sets smaller than 5,000 objects to 100 for the data sets exceeding 15,000 objects. In addition, Spearman's rank correlation coefficient reveals a strong monotonic relation between the speedup and $n$ ($\rho=.72$, $p<.001$). The results of this experiment clearly demonstrate that CCMM not only exhibits a considerably faster performance compared to AMA and SSNAL but also showcases superior scaling properties concerning the size of the analyzed data.

\subsection{Convergence}
\begin{figure}[t]
    \begin{minipage}{\dimexpr0.5\linewidth-1pt}
        \centering
        \includegraphics[scale=1]{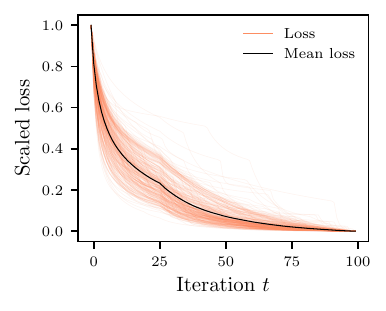}\\[-0.2cm]
        \small{(a) Iterative loss progression in CCMM}
    \end{minipage}
    \begin{minipage}{\dimexpr0.5\linewidth-1pt}
        \centering
        \includegraphics[scale=1]{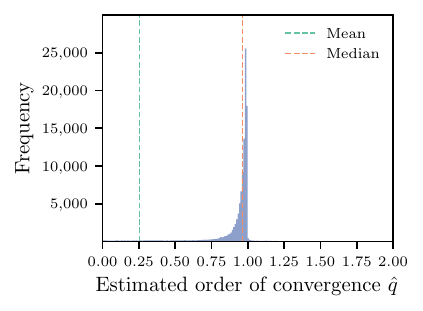}\\[-0.2cm]
        \small{(b) Distribution of the order of convergence}
    \end{minipage}
    \caption{
    In Panel (a): progression of the loss function against the number of iterations for the minimizations by CCMM. Each curve represents a sequence, scaled to the $[0, 1]$ interval, with the mean displayed for clarity. Accelerated convergence is observed after the 26\textsuperscript{th} iteration, representing the burn-in phase. In Panel (b): histogram depicting the distribution of the estimated order of convergence ($\hat{q}$) for CCMM. The focus is on the range from 0 to 2, as approximately 76.7\% of all estimations fall within this interval. The mean value is affected by a few extreme outliers. Of all estimates, 92.8\% are less than 1, indicating sublinear convergence in practice.}
    \label{fig:convergence}
\end{figure}

In this section, we examine the convergence of the CCMM algorithm by analyzing the results from applying CCMM to the two interlocking half-moons data sets with sample sizes ranging from $n=$ 1,000 to $n=$ 5,000 (see Section~\ref{sec:twohalfmooncomparison}).
Figure~\figref{fig:convergence}{a} depicts the progression of the loss function against the number of iterations for the minimizations requiring a minimum of 100 iterations to converge. For clarity, we applied min-max normalization, scaling each sequence to the $[0, 1]$ interval, and display the mean alongside individual curves. The figure highlights the effectiveness of step-doubling, with accelerated convergence observed after the 26\textsuperscript{th} iteration, representing the burn-in phase.

As mentioned in Section~\ref{sec:convergence}, we can gauge the order of convergence using \eqref{eq:orderofconvergence}. For this analysis, we again used the results from the minimizations during the comparison of CCMM, SSNAL, and AMA. Sequences with less than four iterates were excluded since this is the minimum requirement to obtain an estimate for the order of convergence, denoted by $\hat{q}$.
The histogram in Figure~\figref{fig:convergence}{b} provides a visualization of the distribution of $\hat{q}$, focusing on the range from 0 to 2. This zoomed-in view is chosen due to the presence of some extreme outliers that also affect the mean value. Approximately 76.7\% of all estimations fall within the $[0, 2]$ interval, signifying that the majority of the estimates conform to this range. Furthermore, a substantial 92.8\% of estimates are less than 1, indicating that CCMM exhibits sublinear convergence in practical scenarios.

\subsection{Analysis of a Large Data Set}
In prior sections, we have established CCMM as the most competitive algorithm to perform large-scale convex clustering. Here, we showcase CCMM analyzing a data set containing more than one million measurements of the power consumption by a single household located in Sceaux---a commune part of the Paris metropolitan area\footnote{The 
\textit{individual household electric power consumption} data set is made available in the UCI Machine Learning Repository \citep{dua2019uci}.}.

Data consist of measurements sampled at one-minute intervals from the 16\textsuperscript{th} of December 2006 through 16\textsuperscript{th} of December 2008. Metrics include the global active and reactive power (in kilowatts), voltage (in volts), global intensity (in amperes), and the activity at three sub-metering stations (in watt-hours). The first sub-metering station monitored a kitchen, the second a laundry room, and the third a water-heater and air-conditioner. In total, there are 1,048,570 measurements, and the mean of each variable is reported in the first column of Table~\ref{tab:power_consumption}. Before performing the cluster analysis, each variable was standardized to have a mean of zero and variance of one. The weight matrix was computed using $k=15$ and $\phi=0.5$ in combination with the symmetric circulant matrix to ensure a complete cluster hierarchy. 

\begin{table}
\centering
\caption{Variable means for the individual household electric power consumption data for the complete data set (in the leftmost column) and for the two clusters (A and B) that were found using convex clustering.}
\label{tab:power_consumption}
\begin{tabular}{l c rrrrrrrrr}
\toprule
&& Complete & Cluster A & Cluster B \\
\midrule
Global active power 	 && 1.11 	 & 0.62 	 & 2.15 \\
Global reactive power 	 && 0.12 	 & 0.11 	 & 0.13 \\ 
Voltage 	 && 239.97 	 & 240.65 	 & 238.52 \\ 
Global intensity 	 && 4.72 	 & 2.68 	 & 9.05 \\ 
Energy sub-metering No. 1 	 && 1.18 	 & 0.63 	 & 2.33 \\ 
Energy sub-metering No. 2 	 && 1.47 	 & 0.97 	 & 2.54 \\ 
Energy sub-metering No. 3 	 && 5.94 	 & 0.39 	 & 17.71 \\
\midrule
Number of measurements && 1,048,570 	 & 712,749 	 & 335,821 \\
\bottomrule
\end{tabular}
\end{table}

For this analysis, we took an approach distinct from computing the clusterpath for a predetermined sequence for $\lambda$. While useful for comparing different algorithms, it demands advance knowledge of which value of $\lambda$ maps to a specific cluster number output. For this analysis, we thus initialized $\lambda$ at 0.01 and repeatedly minimized the loss function, increasing $\lambda$ by 2.5\% after solving each minimization. As with computing the clusterpath, the solution for the minimization problem at the prior value of $\lambda$ was used as a warm start for the next value of $\lambda$. After reaching 20 clusters, the procedure entered a refinement phase. If the number of clusters was reduced by more than one as $\lambda$ increased, a midpoint value was chosen to obtain a cluster hierarchy as detailed as possible. The procedure was terminated after reaching two clusters since the final reduction in cluster number is trivial. In total, 957 instances of the convex-clustering loss function were solved. The entire procedure was completed in 13.5 hours, thus solving each instance in 51 seconds, on average. To our knowledge, this is the largest data set analyzed to date via convex clustering.

\begin{figure}
    \begin{minipage}{\dimexpr0.5\linewidth-1pt}
        \centering
        \includegraphics[scale=1]{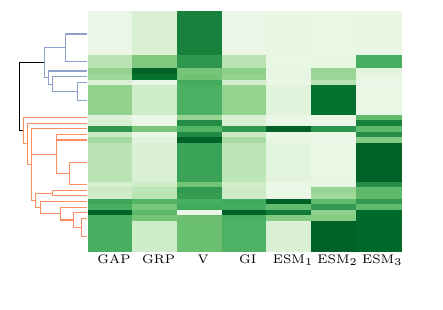}\\[-0.2cm]
        \small{(a) Cluster heatmap}
    \end{minipage}
    \begin{minipage}{\dimexpr0.5\linewidth-1pt}
        \centering
        \includegraphics[scale=1]{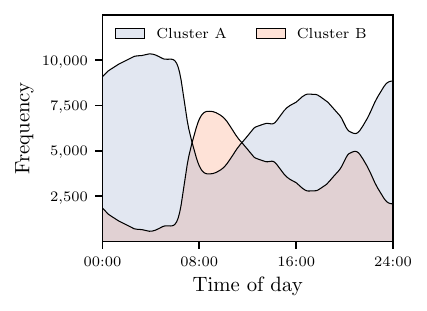}\\[-0.2cm]
        \small{(b) Distribution of the final two clusters}
    \end{minipage}
    \caption{In Panel (a): a cluster heatmap for the final 20 clusters in the individual household electric power consumption data. The height of each row is proportional to the logarithm of the number of measurements in each of the 20 clusters. Cell color is determined by the mean value of the cluster for that particular variable. In Panel (b): the distribution of timestamps in the final two clusters. Frequency is determined by the number of times each timestamp occurred in one of the clusters.}
    \label{fig:hist_heatmap}
\end{figure}

In Figure~\figref{fig:hist_heatmap}{a}, a dendrogram in combination with a cluster heatmap shows the agglomeration of the final 20 clusters. Since these clusters are of different sizes, the height of each cell row in the heatmap is proportional to the logarithm of the size of the corresponding cluster. Of the 20 clusters, there are 15 that contain two or fewer measurements, and the largest two combined comprise over 95\% of the data. The heatmap distinguishes these final two clusters by the measurements at the third energy sub-metering station. Looking at the descriptive statistics of the final two clusters in the second and third columns of Table~\ref{tab:power_consumption}, we can conclude that cluster A corresponds to measurements of lower power consumption, B to greater consumption.

To gain further insight into the final two clusters, Figure~\figref{fig:hist_heatmap}{b} shows the timestamp frequency for both clusters. Energy consumption is generally low during nighttime when members of the household are likely sleeping. Around 08:00 and 20:00, peaks emerge for cluster B to indicate general times of day that people are home, activating the heater or air-conditioner, or performing tasks like cooking or laundering. It should be noted that household members may stick to different time schedules in their power usage activities during weekends, holidays, and changing seasons. Still, this figure conveys a dense level of information about how this household generally spends its days.

\section{Conclusion}
\label{sec:conclusion}
In this paper, we proposed a new, efficient algorithm to minimize the convex clustering loss function and compute the clusterpath. We suggested using a connected, sparse weight matrix in combination with cluster fusions to guarantee a complete cluster hierarchy. At the heart of our minimization algorithm, which we call CCMM, lies an efficient update derived using majorization-minimization.

Our simulations showed that CCMM is up to 10,023 times faster than AMA and up to 1,600 times faster than SSNAL while losing, at most, 0.0008\% accuracy. Furthermore, we analyzed a data set featuring over one million measurements of the electric power consumption by a single household. On average, solving an instance of the convex clustering loss function for a particular value of $\lambda$ took 51 seconds. To our knowledge, this data set is the largest ever clustered using convex clustering.

One useful future research path may merge CCMM and \textit{convex clustering via algorithmic regularization paths} (CARP). Originally developed by \cite{allen2020carp}, CARP uses small increments in the regularization parameter to compute a highly detailed clusterpath. For each value of $\lambda$, its prior solution is used as a warm start, and an ADMM algorithm \citep{boyd2011admm} is allowed to perform one iteration. In their paper, the authors showed that if the increments in $\lambda$ are small enough, the difference between the results of CARP and the exact solution approaches zero. 
Given the efficiency of CCMM, we expect that this could suitably replace the ADMM update in CARP.
Furthermore, we believe it may be possible to derive an algorithm similar to CCMM for convex biclustering \citep{chi2017cvxbiclust}. However, as this is not trivial, these derivations are beyond the scope of the present paper.

In sum, CCMM is an algorithm scaling better than state-of-the-art implementations designed to perform convex clustering. Its excellent scalability facilitates the analysis of larger data sets in less time than possible to date.

\appendix
\section{}
\label{app:fusions}

As detailed in Section~\ref{sec:cluster_fusions}, the potential reduction in computational complexity when minimizing the convex clustering loss function due to cluster fusions can be considerable. Proposition~\ref{theorem:fusions} states that the change in the loss due to a fusion approaches zero if $\varepsilon_f$ is chosen sufficiently small. This appendix contains the proof of Proposition~\ref{theorem:fusions}.\medskip

\noindent
{\bf Proposition \ref{theorem:fusions}}
{\it Let the matrix $\mathbf{M}$ contain the $c$ unique rows of $\mathbf{A}$, and let $\varepsilon_f$ be the threshold used for fusing rows of $\mathbf{M}$. 
Assume that $\| \mathbf{m}_k - \mathbf{m}_l \| \leq \varepsilon_f$ for some $k\neq l$. 
Let $\mathbf{m}_{new}$ be the weighted average of $\mathbf{m}_k$ and $\mathbf{m}_l$ computed as in \eqref{eq:fusions_weighted_mean}, and let $\mathbf{M}_{new}$ be the matrix that results from setting $\mathbf{m}_k$ and $\mathbf{m}_l$ to $\mathbf{m}_{new}$. Then, for fixed $\lambda$ and $\mathbf{W}$, the absolute error $|L(\M) - L(\M_{new})|$ tends toward zero as $\varepsilon_f \rightarrow 0$.} \medskip

\begin{proof}
If $\| \mathbf{m}_k - \mathbf{m}_l \| \leq \varepsilon_f$, the new cluster centroid is computed as in \eqref{eq:fusions_weighted_mean}. Let
\begin{align*}
    \bm{\sigma}_k &= \mathbf{m}_k - \mathbf{m}_{new} = \frac{|C_l|}{|C_k| + |C_l|} (\mathbf{m}_k - \mathbf{m}_l) \\
    \bm{\sigma}_l &= \mathbf{m}_l - \mathbf{m}_{new} = \frac{|C_k|}{|C_k| + |C_l|} (\mathbf{m}_l - \mathbf{m}_k),
\end{align*}
such that
\begin{equation}
\label{eq:sigma_bound}
    \| \bm{\sigma}_k \| \leq \frac{|C_l|}{|C_k| + |C_l|} \varepsilon_f \quad \text{and} \quad \| \bm{\sigma}_l \| \leq \frac{|C_k|}{|C_k| + |C_l|} \varepsilon_f.
\end{equation}
Writing out the absolute value of the difference between the loss before and after merging $\mathbf{m}_k$ and $\mathbf{m}_l$, we obtain
\begin{align*}
    |L(\M) - L(\M_{new})| \opl{=} \Biggl| \KE \Biggl( \sum_{i \in C_k} \| \mathbf{x}_i - \mathbf{m}_k \|^2 + \sum_{i \in C_l} \| \mathbf{x}_i - \mathbf{m}_l \|^2 \\
    \opr{-} \sum_{i \in C_k} \| \mathbf{x}_i - \mathbf{m}_{new} \|^2 - \sum_{i \in C_l} \| \mathbf{x}_i - \mathbf{m}_{new} \|^2 \Biggr)  \\ 
    \opr{+} \lambda\KP \Biggl( \sum_{\substack{i=0 \\ i \notin \{k, l\}}}^c
    % (\UWU)_{ki}
    \uWu{k}{i}
    \| \mathbf{m}_k - \mathbf{m}_i \| \\
    \opr{+} \sum_{\substack{i=0 \\ i \notin \{k, l\}}}^c 
    % (\UWU)_{li}
    \uWu{l}{i}
    \| \mathbf{m}_l - \mathbf{m}_i \| + 
    % (\UWU)_{kl}
    \uWu{k}{l}
    \| \mathbf{m}_k - \mathbf{m}_l \| \\
    \opr{-} \sum_{\substack{i=0 \\ i \notin \{k, l\}}}^c
    % (\UWU)_{ki}
    \uWu{k}{i}
    \| \mathbf{m}_{new} - \mathbf{m}_i \| - \sum_{\substack{i=0 \\ i \notin \{k, l\}}}^c
    % (\UWU)_{li}
    \uWu{l}{i}
    \| \mathbf{m}_{new} - \mathbf{m}_i \| \Biggr) \Biggr|,
\end{align*}
where the first two lines concern the difference between the first terms of $L(\M)$ and $L(\M_{new})$ and the remaining three lines concern the difference between the values of the penalty terms. Using the triangle inequality, this can be written as
\begin{equation*}
    |L(\M) - L(\M_{new})| \leq \KE | \Gamma_1 | + \lambda \KP |\Gamma_2|,
\end{equation*}
where $\Gamma_1$ is the result of the terms on the first two lines and $\Gamma_2$ is the result of the terms on the remaining four lines. We start by providing an upper bound for $|\Gamma_1|$. Rewriting yields
\begin{align*}
    |\Gamma_1| &= \Biggl| \sum_{i \in C_k} \bigl( \| \mathbf{x}_i - \mathbf{m}_k \|^2 - \| \mathbf{x}_i - \mathbf{m}_{new} \|^2 \bigr) + \sum_{i \in C_l} \bigl( \| \mathbf{x}_i - \mathbf{m}_l \|^2 - \| \mathbf{x}_i - \mathbf{m}_{new} \|^2 \bigr) \Biggr| \\
    &= \Biggl| \sum_{i \in C_k} \bigl( \| \mathbf{x}_i - \mathbf{m}_k \|^2 - \| \mathbf{x}_i - \mathbf{m}_{k} + \bm{\sigma}_k \|^2 \bigr) + \sum_{i \in C_l} \bigl( \| \mathbf{x}_i - \mathbf{m}_l \|^2 - \| \mathbf{x}_i - \mathbf{m}_{l} + \bm{\sigma}_l \|^2 \bigr) \Biggr| \\
    &= \Biggl| \sum_{i \in C_k} \bigl( \bm{\sigma}_k\T\bm{\sigma}_k - 2(\mathbf{x}_i - \mathbf{m}_k)\T \bm{\sigma}_k \bigr) + \sum_{i \in C_l} \bigl( \bm{\sigma}_l\T\bm{\sigma}_l - 2(\mathbf{x}_i - \mathbf{m}_l)\T \bm{\sigma}_l \bigr) \Biggr|,
\end{align*}
which is bounded by
\begin{align*}
    |\Gamma_1| &\leq \sum_{i \in C_k} \bigl( \bm{\sigma}_k\T\bm{\sigma}_k + 2|(\mathbf{x}_i - \mathbf{m}_k)\T \bm{\sigma}_k | \bigr) + \sum_{i \in C_l} \bigl( \bm{\sigma}_l\T\bm{\sigma}_l + 2|(\mathbf{x}_i - \mathbf{m}_l)\T \bm{\sigma}_l | \bigr) \\
    &\leq \sum_{i \in C_k} \bigl( \bm{\sigma}_k\T\bm{\sigma}_k + 2 \| \mathbf{x}_i - \mathbf{m}_k \| \| \bm{\sigma}_k \| \bigr) + \sum_{i \in C_l} \bigl( \bm{\sigma}_l\T\bm{\sigma}_l + 2 \| \mathbf{x}_i - \mathbf{m}_l \| \| \bm{\sigma}_l \| \bigl) \\
    % &\leq \frac{|C_k|^2 + |C_l|^2}{(|C_k| + |C_l|)^2} 
    &\leq \frac{|C_k| \, |C_l|}{|C_k| + |C_l|} 
    \varepsilon_f^2 + \frac{2}{|C_k| + |C_l|} \Biggl( |C_l| \sum_{i \in C_k} \| \mathbf{x}_i - \mathbf{m}_k \| + |C_k|\sum_{i \in C_l} \| \mathbf{x}_i - \mathbf{m}_l \| \Biggr) \varepsilon_f,
\end{align*}
where we made use of the triangle inequality, the Cauchy-Schwarz inequality, and \eqref{eq:sigma_bound}. 
Each of the terms in the upper bound is now proportional to $\varepsilon_f$. To bound $|\Gamma_2|$, consider
\begin{align*}
    \| \mathbf{m}_k - \mathbf{m}_i \| - \| \mathbf{m}_{new} - \mathbf{m}_i \| &= \| \mathbf{m}_k - \mathbf{m}_i \| - \| \mathbf{m}_k - \bm{\sigma}_k - \mathbf{m}_i \| \\
    &= \| \bm{\sigma}_k + \mathbf{m}_k - \bm{\sigma}_k - \mathbf{m}_i \| - \| \mathbf{m}_k - \bm{\sigma}_k - \mathbf{m}_i \|,
\end{align*}
and by means of the triangle inequality we obtain
\begin{align*}
    \| \bm{\sigma}_k + \mathbf{m}_k - \bm{\sigma}_k - \mathbf{m}_i \| &\leq \| \bm{\sigma}_k \| + \| \mathbf{m}_k - \bm{\sigma}_k - \mathbf{m}_i \| \\
    \| \mathbf{m}_k - \mathbf{m}_i \| - \| \mathbf{m}_k - \bm{\sigma}_k - \mathbf{m}_i \| &\leq \| \bm{\sigma}_k \|.
\end{align*}
This can be combined with
\begin{align*}
    \| \mathbf{m}_k - \bm{\sigma}_k - \mathbf{m}_i \| &\leq \| \mathbf{m}_k - \mathbf{m}_i \| + \| \bm{\sigma}_k \| \\
    \| \mathbf{m}_k - \bm{\sigma}_k - \mathbf{m}_i \| - \| \mathbf{m}_k - \mathbf{m}_i \| &\leq  \| \bm{\sigma}_k \|,
\end{align*}
to obtain
\begin{equation}
\label{eq:bound_theorem_2}
    \Bigl| \| \mathbf{m}_k - \mathbf{m}_i \| - \| \mathbf{m}_{new} - \mathbf{m}_i \| \Bigr| \leq \| \bm{\sigma}_k \|.
\end{equation}
The upper bound for $|\Gamma_2|$ is
\begin{align*}
    | \Gamma_2 | \opl{=} \Biggl| \sum_{\substack{i=0 \\ i \notin \{k, l\}}}^c 
    % (\UWU)_{ki}
    \uWu{k}{i}
    ( \| \mathbf{m}_k - \mathbf{m}_i \| - \| \mathbf{m}_{new} - \mathbf{m}_i \| ) \\
    \opr{+} \sum_{\substack{i=0 \\ i \notin \{k, l\}}}^c
    % (\UWU)_{li}
    \uWu{l}{i}
    ( \| \mathbf{m}_l - \mathbf{m}_i \| - \| \mathbf{m}_{new} - \mathbf{m}_i \| ) + 
    % (\UWU)_{kl}
    \uWu{k}{l}
    \| \mathbf{m}_k - \mathbf{m}_l \| \Biggr| \\
    \opl{\leq} \sum_{\substack{i=0 \\ i \notin \{k, l\}}}^c 
    % (\UWU)_{ki}
    \uWu{k}{i}
    \| \bm{\sigma}_k \| + \sum_{\substack{i=0 \\ i \notin \{k, l\}}}^c 
    % (\UWU)_{li}
    \uWu{l}{i}
    \| \bm{\sigma}_l \| + 
    % (\UWU)_{kl}
    \uWu{k}{l}
    \| \mathbf{m}_k - \mathbf{m}_l \| \\
    \opl{\leq} \Biggl( 
    % (\UWU)_{kl}
    \uWu{k}{l}
    + \frac{|C_l|}{|C_k| + |C_l|} \sum_{\substack{i=0 \\ i \notin \{k, l\}}}^c 
    % (\UWU)_{ki}
    \uWu{k}{i}
    + \frac{|C_k|}{|C_k| + |C_l|} \sum_{\substack{i=0 \\ i \notin \{k, l\}}}^c
    % (\UWU)_{li}
    \uWu{l}{i}
    \Biggr) \varepsilon_f,
\end{align*}
where we made use of the triangle inequality, \eqref{eq:sigma_bound}, and \eqref{eq:bound_theorem_2}. From these two upper bounds, we can conclude that if $\varepsilon_f \rightarrow 0$, both $|\Gamma_1| \rightarrow 0$ and $|\Gamma_2| \rightarrow 0$. Hence, $|L(\M) - L(\M_{new})| \rightarrow 0$ as $\varepsilon_f \rightarrow 0$.
\end{proof}

\vskip 0.2in
\bibliography{cvx_clustering}

\begin{thebibliography}{54}
\providecommand{\natexlab}[1]{#1}
\providecommand{\url}[1]{\texttt{#1}}
\expandafter\ifx\csname urlstyle\endcsname\relax
  \providecommand{\doi}[1]{doi: #1}\else
  \providecommand{\doi}{doi: \begingroup \urlstyle{rm}\Url}\fi

\bibitem[Arya et~al.(2019)Arya, Mount, Kemp, and Jefferis]{arya2019knn}
S.~Arya, D.~Mount, S.~E. Kemp, and G.~Jefferis.
\newblock \emph{{RANN: Fast Nearest Neighbour Search (Wraps ANN Library) Using
  L2 Metric}}, 2019.
\newblock URL \url{https://CRAN.R-project.org/package=RANN}.
\newblock R package version 2.6.1.

\bibitem[Bentley(1975)]{bentley1975kdtree}
J.~L. Bentley.
\newblock {Multidimensional Binary Search Trees Used for Associative
  Searching}.
\newblock \emph{Communications of the ACM}, 18\penalty0 (9):\penalty0 509--517,
  1975.

\bibitem[Bolte and Pauwels(2016)]{bolte2016majorization}
J.~Bolte and E.~Pauwels.
\newblock {Majorization-Minimization Procedures and Convergence of SQP Methods
  for Semi-Algebraic and Tame Programs}.
\newblock \emph{Mathematics of Operations Research}, 41\penalty0 (2):\penalty0
  442--465, 2016.

\bibitem[Boyd et~al.(2011)Boyd, Parikh, Chu, Peleato, and
  Eckstein]{boyd2011admm}
S.~P. Boyd, N.~Parikh, E.~Chu, B.~Peleato, and J.~Eckstein.
\newblock {Distributed Optimization and Statistical Learning via the
  Alternating Direction Method of Multipliers}.
\newblock \emph{Foundations and Trends{\textregistered} in Machine learning},
  3\penalty0 (1):\penalty0 1--122, 2011.

\bibitem[Bronson(1989)]{bronson1989gershgorin}
R.~Bronson.
\newblock \emph{{Schaum's Outline of Theory and Problems of Matrix
  Operations}}.
\newblock McGraw-Hill, 1989.

\bibitem[Chen et~al.(2015)Chen, Chi, Ranola, and Lange]{chen2015convex}
G.~K. Chen, E.~C. Chi, J.~M.~O. Ranola, and K.~Lange.
\newblock {Convex Clustering: An Attractive Alternative to Hierarchical
  Clustering}.
\newblock \emph{PLoS Computational Biology}, 11\penalty0 (5):\penalty0 1--31,
  2015.

\bibitem[Chi and Lange(2015)]{chi2015ama}
E.~C. Chi and K.~L. Lange.
\newblock {Splitting Methods for Convex Clustering}.
\newblock \emph{Journal of Computational and Graphical Statistics}, 24\penalty0
  (4):\penalty0 994--1013, 2015.

\bibitem[Chi and Steinerberger(2019)]{chi2019recovering}
E.~C. Chi and S.~Steinerberger.
\newblock {Recovering Trees with Convex Clustering}.
\newblock \emph{SIAM Journal on Mathematics of Data Science}, 1\penalty0
  (3):\penalty0 383--407, 2019.

\bibitem[Chi et~al.(2017)Chi, Allen, and Baraniuk]{chi2017cvxbiclust}
E.~C. Chi, G.~I. Allen, and R.~G. Baraniuk.
\newblock {Convex Biclustering}.
\newblock \emph{Biometrics}, 73\penalty0 (1):\penalty0 10--19, 2017.

\bibitem[Chiquet et~al.(2017)Chiquet, Gutierrez, and Rigaill]{chiquet2017fast}
J.~Chiquet, P.~Gutierrez, and G.~Rigaill.
\newblock {Fast Tree Inference with Weighted Fusion Penalties}.
\newblock \emph{Journal of Computational and Graphical Statistics}, 26\penalty0
  (1):\penalty0 205--216, 2017.

\bibitem[Cordero and Torregrosa(2007)]{cordero2007variants}
A.~Cordero and J.~R. Torregrosa.
\newblock {Variants of Newton’s Method Using Fifth-Order Quadrature
  Formulas}.
\newblock \emph{Applied Mathematics and Computation}, 190\penalty0
  (1):\penalty0 686--698, 2007.

\bibitem[Deng(2012)]{deng2012mnist}
L.~Deng.
\newblock {The MNIST Database of Handwritten Digit Images for Machine Learning
  Research}.
\newblock \emph{IEEE Signal Processing Magazine}, 29\penalty0 (6):\penalty0
  141--142, 2012.

\bibitem[Dua and Graff(2019)]{dua2019uci}
D.~Dua and C.~Graff.
\newblock {UCI Machine Learning Repository}, 2019.
\newblock URL \url{http://archive.ics.uci.edu/ml}.

\bibitem[Eddelbuettel(2013)]{eddel2013rcpp}
D.~Eddelbuettel.
\newblock \emph{{Seamless R and C\texttt{++} Integration with Rcpp}}.
\newblock Springer, 2013.

\bibitem[Eddelbuettel and Balamuta(2017)]{eddel2017rcpp}
D.~Eddelbuettel and J.~J. Balamuta.
\newblock {Extending R with C\texttt{++}: A Brief Introduction to Rcpp}.
\newblock \emph{PeerJ Preprints}, 5:\penalty0 e3188v1, 2017.

\bibitem[Eddelbuettel and Fran\c{c}ois(2011)]{eddel2011rcpp}
D.~Eddelbuettel and R.~Fran\c{c}ois.
\newblock {Rcpp: Seamless R and C\texttt{++} Integration}.
\newblock \emph{Journal of Statistical Software}, 40\penalty0 (8):\penalty0
  1--18, 2011.

\bibitem[Fodor et~al.(2022)Fodor, Jakoveti{\'c}, Boberi{\'c}~Krsti{\'c}ev, and
  {\v{S}}krbi{\'c}]{fodor2022parallel}
L.~Fodor, D.~Jakoveti{\'c}, D.~Boberi{\'c}~Krsti{\'c}ev, and
  S.~{\v{S}}krbi{\'c}.
\newblock {A Parallel ADMM-based Convex Clustering Method}.
\newblock \emph{EURASIP Journal on Advances in Signal Processing},
  2022\penalty0 (1):\penalty0 1--33, 2022.

\bibitem[Gan et~al.(2007)Gan, Ma, and Wu]{ma2007clustering_book}
G.~Gan, C.~Ma, and J.~Wu.
\newblock \emph{Data Clustering: Theory, Algorithms, and Applications}.
\newblock SIAM, 2007.

\bibitem[Gower and Groenen(1991)]{gower1990symm_circ}
J.~C. Gower and P.~J.~F. Groenen.
\newblock {Applications of the Modified Leverrier-Faddeev Algorithm for the
  Construction of Explicit Matrix Spectral Decompositions and Inverses}.
\newblock \emph{Utilitas Mathematica}, 40:\penalty0 51--64, 1991.

\bibitem[Groenen et~al.(1999)Groenen, Heiser, and
  Meulman]{groenen1999diag_major}
P.~J.~F. Groenen, W.~J. Heiser, and J.~J. Meulman.
\newblock {Global Optimization in Least Squares Multidimensional Scaling by
  Distance Smoothing}.
\newblock \emph{Journal of Classification}, 16\penalty0 (2):\penalty0 225--254,
  1999.

\bibitem[Guennebaud and Jacob(2010)]{eigen}
G.~Guennebaud and B.~Jacob.
\newblock Eigen v3.
\newblock http://eigen.tuxfamily.org, 2010.

\bibitem[Guo et~al.(2010)Guo, Levina, Michailidis, and Zhu]{guo2010pairwise}
J.~Guo, E.~Levina, G.~Michailidis, and J.~Zhu.
\newblock {Pairwise Variable Selection for High-dimensional Model-based
  Clustering}.
\newblock \emph{Biometrics}, 66\penalty0 (3):\penalty0 793--804, 2010.

\bibitem[Havel(1991)]{havel1991mm_convergence}
T.~F. Havel.
\newblock {An Evaluation of Computational Strategies for Use in the
  Determination of Protein Structure from Distance Constrains Obtained by
  Nuclear Magnetic Resonance}.
\newblock \emph{Progress in Biophysics and Molecular Research}, 56\penalty0
  (1):\penalty0 43--78, 1991.

\bibitem[Hocking et~al.(2011)Hocking, Joulin, Bach, and
  Vert]{hocking2011clusterpath}
T.~D. Hocking, A.~Joulin, F.~Bach, and J.-P. Vert.
\newblock {Clusterpath: An Algorithm for Clustering Using Convex Fusion
  Penalties}.
\newblock In \emph{{The 28th International Conference on Machine Learning}},
  Bellevue, Washington, 2011.

\bibitem[Hunter and Lange(2004)]{hunter2004IM_tutorial}
D.~R. Hunter and K.~L. Lange.
\newblock {A Tutorial on MM Algorithms}.
\newblock \emph{The American Statistician}, 58\penalty0 (1):\penalty0 30--37,
  2004.

\bibitem[Jakob et~al.(2017)Jakob, Rhinelander, and Moldovan]{pybind11}
W.~Jakob, J.~Rhinelander, and D.~Moldovan.
\newblock {pybind11---Seamless operability between C\texttt{++}11 and Python},
  2017.
\newblock https://github.com/pybind/pybind11.

\bibitem[Kruskal(1956)]{kruskal1956mst}
J.~B. Kruskal.
\newblock {On the Shortest Spanning Subtree of a Graph and the Traveling
  Salesman Problem}.
\newblock \emph{Proceedings of the American Mathematical society}, 7\penalty0
  (1):\penalty0 48--50, 1956.

\bibitem[Lange(2013)]{lange2013optimization}
K.~Lange.
\newblock \emph{Optimization}, volume~95.
\newblock Springer Science \& Business Media, 2013.

\bibitem[Lange et~al.(2000)Lange, Hunter, and Yang]{lange2000IM_optimization}
K.~L. Lange, D.~R. Hunter, and I.~Yang.
\newblock {Optimization Transfer Using Surrogate Objective Functions}.
\newblock \emph{Journal of Computational and Graphical Statistics}, 9\penalty0
  (1):\penalty0 1--20, 2000.

\bibitem[{\DE{Leeuw}{de}{de}}~Leeuw(1977)]{leeuw1977IM}
J.~{\DE{Leeuw}{de}{de}}~Leeuw.
\newblock {Applications of Convex Analysis to Multidimensional Scaling}.
\newblock In J.-R. Barra, F.~cois Brodeau, G.~Romier, and B.~van Cutsem,
  editors, \emph{Recent Developments in Statistics}, pages 133--146. North
  Holland Publishing Company, 1977.

\bibitem[{\DE{Leeuw}{de}{de}}~Leeuw(1988)]{leeuw1988convergence}
J.~{\DE{Leeuw}{de}{de}}~Leeuw.
\newblock {Convergence of the Majorization Method for Multidimensional
  Scaling}.
\newblock \emph{Journal of classification}, 5:\penalty0 163--180, 1988.

\bibitem[{\DE{Leeuw}{de}{de}}~Leeuw(1994)]{leeuw1994mm_convergence}
J.~{\DE{Leeuw}{de}{de}}~Leeuw.
\newblock {Block-Relaxation Algorithms in Statistics}.
\newblock In H.-H. Bock, W.~Lenski, and M.~M. Richter, editors,
  \emph{Information Systems and Data Analysis}, pages 308--324. Springer Berlin
  Heidelberg, 1994.

\bibitem[{\DE{Leeuw}{de}{de}}~Leeuw and Heiser(1980)]{deleeuw1980step_doubling}
J.~{\DE{Leeuw}{de}{de}}~Leeuw and W.~J. Heiser.
\newblock {Multidimensional Scaling with Restrictions on the Configuration}.
\newblock \emph{Multivariate Analysis}, 5\penalty0 (1):\penalty0 501--522,
  1980.

\bibitem[Lindsten et~al.(2011)Lindsten, Ohlsson, and
  Ljung]{lindsten2011convex_also}
F.~Lindsten, H.~Ohlsson, and L.~Ljung.
\newblock {Just Relax and Come Clustering!: A Convexification of K-Means
  Clustering}.
\newblock Technical report, Department of Electrical Engineering, Link{\"o}ping
  University, Link{\"o}ping, Sweden, 2011.

\bibitem[MacQueen(1967)]{macqueen1967kmeans}
J.~MacQueen.
\newblock {Some Methods for Classification and Analysis of Multivariate
  Observations}.
\newblock In \emph{Proceedings of the Fifth Berkeley Symposium on Mathematical
  Statistics and Probability}, pages 281--297, 1967.

\bibitem[Mairal(2013)]{mairal2013stochastic}
J.~Mairal.
\newblock {Stochastic Majorization-Minimization Algorithms for Large-scale
  Optimization}.
\newblock \emph{Advances in Neural Information Processing Systems}, 26, 2013.

\bibitem[Mairal(2015)]{mairal2015incremental}
J.~Mairal.
\newblock {Incremental Majorization-Minimization Optimization with Application
  to Large-scale Machine Learning}.
\newblock \emph{SIAM Journal on Optimization}, 25\penalty0 (2):\penalty0
  829--855, 2015.

\bibitem[Marchetti and Zhou(2014)]{marchetti2014solution}
Y.~Marchetti and Q.~Zhou.
\newblock {Solution Path Clustering with Adaptive Concave Penalty}.
\newblock \emph{Electronic Journal of Statistics}, 8\penalty0 (1):\penalty0
  1569 -- 1603, 2014.

\bibitem[McInnes et~al.(2018)McInnes, Healy, and Melville]{mcinnes2018umap}
L.~McInnes, J.~Healy, and J.~Melville.
\newblock {UMAP: Uniform Manifold Approximation and Projection for Dimension
  Reduction}, 2018.
\newblock URL \url{https://arxiv.org/abs/1802.03426}.

\bibitem[Ortega and Rheinboldt(1970)]{ortega1970iterative}
J.~M. Ortega and W.~C. Rheinboldt.
\newblock \emph{{Iterative Solution of Nonlinear Equations in Several
  Variables}}.
\newblock Academic Press, 1970.

\bibitem[Pedregosa et~al.(2011)Pedregosa, Varoquaux, Gramfort, Michel, Thirion,
  Grisel, Blondel, Prettenhofer, Weiss, Dubourg, Vanderplas, Passos,
  Cournapeau, Brucher, Perrot, and Duchesnay]{scikit-learn}
F.~Pedregosa, G.~Varoquaux, A.~Gramfort, V.~Michel, B.~Thirion, O.~Grisel,
  M.~Blondel, P.~Prettenhofer, R.~Weiss, V.~Dubourg, J.~Vanderplas, A.~Passos,
  D.~Cournapeau, M.~Brucher, M.~Perrot, and E.~Duchesnay.
\newblock {Scikit-learn: Machine Learning in Python}.
\newblock \emph{Journal of Machine Learning Research}, 12:\penalty0 2825--2830,
  2011.

\bibitem[Pelckmans et~al.(2005)Pelckmans, de~Brabanter, Suykens, and
  de~Moor]{pelckmans2005convex_first}
K.~Pelckmans, J.~de~Brabanter, J.~A.~K. Suykens, and B.~de~Moor.
\newblock Convex {C}lustering {S}hrinkage.
\newblock In \emph{PASCAL Workshop on Statistics and Optimization of Clustering
  Workshop}, 2005.

\bibitem[Radchenko and Mukherjee(2017)]{radchenko2017cvx}
P.~Radchenko and G.~Mukherjee.
\newblock {Convex Clustering via $\ell_1$ Fusion Penalization}.
\newblock \emph{Journal of the Royal Statistical Society Series B: Statistical
  Methodology}, 79\penalty0 (5):\penalty0 1527--1546, 2017.

\bibitem[Razaviyayn et~al.(2013)Razaviyayn, Hong, and
  Luo]{razaviyayn2013unified}
M.~Razaviyayn, M.~Hong, and Z.-Q. Luo.
\newblock {A Unified Convergence Analysis of Block Successive Minimization
  Methods for Nonsmooth Optimization}.
\newblock \emph{SIAM Journal on Optimization}, 23\penalty0 (2):\penalty0
  1126--1153, 2013.

\bibitem[She(2008)]{she2008sparse}
Y.~She.
\newblock \emph{{Sparse Regression with Exact Clustering}}.
\newblock Stanford University, 2008.

\bibitem[Sun et~al.(2021)Sun, Toh, and Yuan]{sun2018sparse_convex}
D.~Sun, K.-C. Toh, and Y.~Yuan.
\newblock {Convex Clustering: Model, Theoretical Guarantee and Efficient
  Algorithm}.
\newblock \emph{Journal of Machine Learning Research}, 22\penalty0
  (9):\penalty0 1--32, 2021.

\bibitem[Vo{\ss} and Eckhardt(1980)]{voss1980IM}
H.~Vo{\ss} and U.~Eckhardt.
\newblock {Linear Convergence of Generalized Weiszfeld's Method}.
\newblock \emph{Computing}, 25\penalty0 (3):\penalty0 243--251, 1980.

\bibitem[Weiszfeld(1937)]{weisz1937im}
E.~Weiszfeld.
\newblock {Sur le Point pour lequel la Somme des Distances de $n$ Points
  Donn\'es Est Minimum}.
\newblock \emph{Tohoku Mathemacial Journal}, 43:\penalty0 355--386, 1937.

\bibitem[Weylandt et~al.(2020)Weylandt, Nagorski, and Allen]{allen2020carp}
M.~Weylandt, J.~Nagorski, and G.~I. Allen.
\newblock {Dynamic Visualization and Fast Computation for Convex Clustering via
  Algorithmic Regularization}.
\newblock \emph{Journal of Computational and Graphical Statistics}, 29\penalty0
  (1):\penalty0 87--96, 2020.

\bibitem[Xu et~al.(2016)Xu, Lin, Zhao, and Zha]{xu2016relaxed}
C.~Xu, Z.~Lin, Z.~Zhao, and H.~Zha.
\newblock {Relaxed Majorization-Minimization for Non-smooth and Non-convex
  Optimization}.
\newblock In \emph{Proceedings of the AAAI Conference on Artificial
  Intelligence}, volume~30, 2016.

\bibitem[Yi et~al.(2021)Yi, Huang, Mishne, and Chi]{yi2021cobrac}
H.~Yi, L.~Huang, G.~Mishne, and E.~C. Chi.
\newblock {COBRAC: A Fast Implementation of Convex Biclustering with
  Compression}.
\newblock \emph{Bioinformatics}, 37\penalty0 (20):\penalty0 3667--3669, 2021.

\bibitem[Yuan and Lin(2006)]{yuan2006grouped_lasso}
M.~Yuan and Y.~Lin.
\newblock {Model Selection and Estimation in Regression with Grouped
  Variables}.
\newblock \emph{Journal of the Royal Statistical Society: Series B (Statistical
  Methodology)}, 68\penalty0 (1):\penalty0 49--67, 2006.

\bibitem[Yuille and Rangarajan(2003)]{yuille2003cccp}
A.~L. Yuille and A.~Rangarajan.
\newblock {The Concave-Convex Procedure}.
\newblock \emph{Neural Computation}, 15\penalty0 (4):\penalty0 915--936, 2003.

\bibitem[Zhou et~al.(2021)Zhou, Yi, Mishne, and Chi]{zhou2021scalable}
W.~Zhou, H.~Yi, G.~Mishne, and E.~Chi.
\newblock {Scalable algorithms for convex clustering}.
\newblock In \emph{2021 IEEE Data Science and Learning Workshop (DSLW)}, pages
  1--6, 2021.

\end{thebibliography}

\end{document}